\documentclass[11pt,a4paper]{article}
\usepackage[dvipsnames, svgnames, x11names]{xcolor}
\usepackage[hyperref]{acl2020}
\usepackage{amsmath,amsfonts,bm}
\def\eqref#1{equation~\ref{#1}}
\def\1{\bm{1}}

\DeclareMathAlphabet{\mathsfit}{\encodingdefault}{\sfdefault}{m}{sl}
\SetMathAlphabet{\mathsfit}{bold}{\encodingdefault}{\sfdefault}{bx}{n}
\newcommand{\tens}[1]{\bm{\mathsfit{#1}}}

\def\tE{{\tens{E}}}

\usepackage{times}
\usepackage{latexsym}

\usepackage{url}
\usepackage{listings}
\usepackage{algorithmic}
\usepackage{algorithm}
\newtheorem{theorem}{Theorem}
\newtheorem{definition}{Definition}
\usepackage{graphicx}
\usepackage{microtype}

\usepackage{multirow}

\usepackage{booktabs}
\usepackage{amssymb}
\usepackage{amsmath}
\usepackage{enumitem}
\usepackage{mathtools}
\usepackage{fdsymbol}
\usepackage{amsmath,amsfonts,bm}
\newcommand{\nop}[1]{}
\newcommand{\eg}{{\sl e.g.}}
\newcommand{\ie}{{\sl i.e.}}

\usepackage{fontawesome} 
\usepackage{pifont}
\usepackage{footnote}
\newcommand{\sectionref}[1]{\S\ref{#1}}
\newcommand{\cmark}{\ding{51}}%
\newcommand{\xmark}{\ding{55}}%

\makesavenoteenv{tabular}
\makesavenoteenv{table}
 \aclfinalcopy % (by Xiang: uncomment for now) Uncomment this line for the final submission
\def\aclpaperid{1648} %  Enter the acl Paper ID here

\title{Scalable Multi-Hop Relational Reasoning for\\ Knowledge-Aware Question Answering}

\author{
 Yanlin Feng\textsuperscript{{$\clubsuit$}}\thanks{~~{The first two authors contributed equally. The major work was done when both authors interned at USC.}}~~
 Xinyue Chen\textsuperscript{$\spadesuit$}$^*$~
 Bill Yuchen Lin\textsuperscript{$\varheartsuit$}~
 Peifeng Wang\textsuperscript{$\varheartsuit$}~
 Jun Yan\textsuperscript{$\varheartsuit$}~
 \textbf{Xiang Ren}\textsuperscript{$\varheartsuit$}\\
 \texttt{fengyanlin@pku.edu.cn, xinyuech@andrew.cmu.edu,}
 \\
 \texttt{\{yuchen.lin, peifengw, yanjun, xiangren\}@usc.edu} \\ 
 \textsuperscript{$\varheartsuit$}{University of Southern California} \\ \textsuperscript{$\clubsuit$}Peking University \quad \textsuperscript{$\spadesuit$}Carnegie Mellon University 
}

\begin{document}
\maketitle

\begin{abstract}
% While fine-tuning pre-trained language models (PTLMs) has yielded strong results on a range of question answering (QA) benchmarks, these methods still suffer in cases when external knowledge is needed to infer the right answer. 
% Existing work on augmenting QA models with external knowledge (e.g., knowledge graphs) either struggle to model multi-hop relations efficiently, or lack transparency into the model's prediction rationale. 
Existing work on augmenting question answering (QA) models with external knowledge (e.g., knowledge graphs) either struggle to model multi-hop relations efficiently, or lack transparency into the model's prediction rationale. 
In this paper, we propose a novel knowledge-aware approach that equips pre-trained language models (PTLMs) with a \textit{multi-hop relational reasoning module}, named multi-hop graph relation network~(MHGRN).
It performs multi-hop, multi-relational reasoning over subgraphs extracted from external knowledge graphs. 
The proposed reasoning module unifies path-based reasoning methods and graph neural networks to achieve better interpretability and scalability. 
We also empirically show its effectiveness and scalability on CommonsenseQA and OpenbookQA datasets, and interpret its behaviors with case studies\footnote{\url{https://github.com/INK-USC/MHGRN}}.
% In particular, MHGRN is now only second to UnifiedQA~\cite{khashabi2020unifiedqa}, a large model based on T5~\cite{t5} on the CommonsenseQA official test set.
% \footnote{Our code is submitted and will be public after the review.}
\end{abstract}

\section{Introduction}
\label{sec:intro}
\begin{figure}[t]
%  \vspace{-0.1cm}
      \centering
        \includegraphics[width=0.85\linewidth]{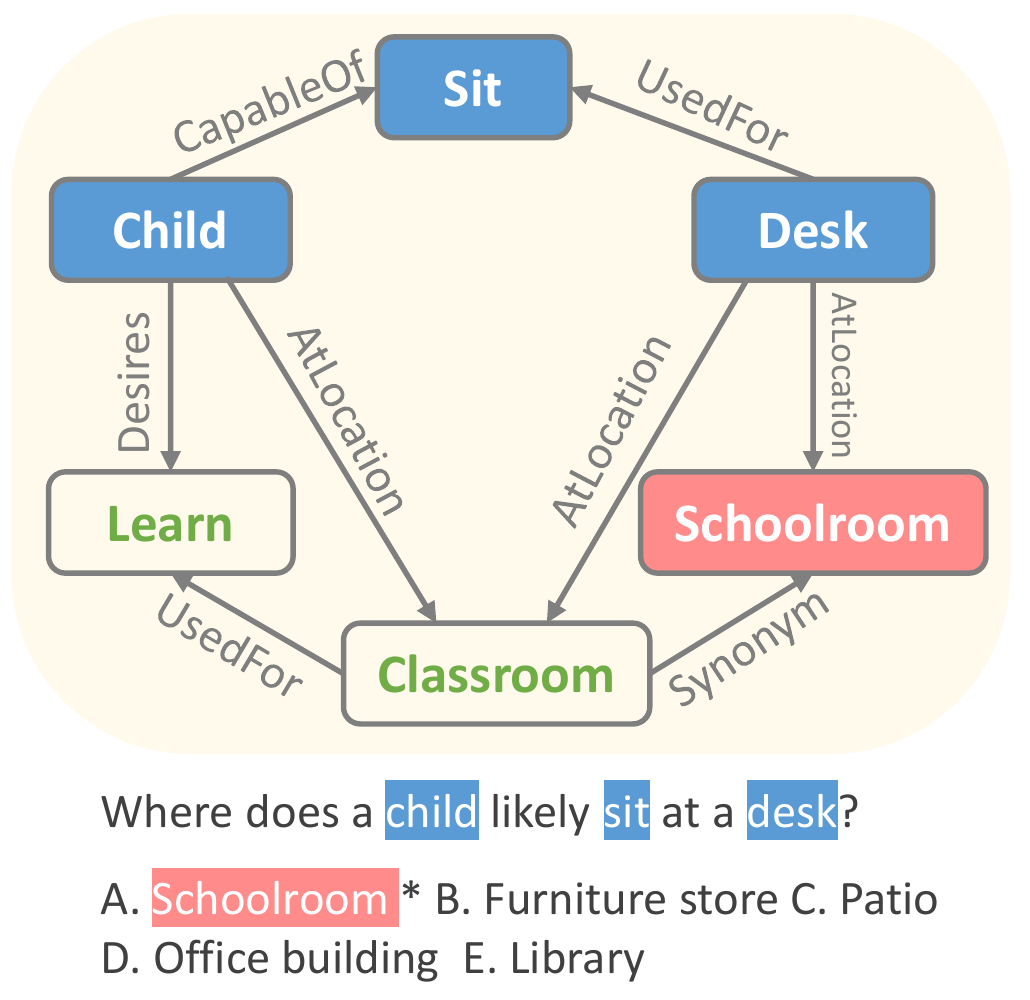}
        % \vspace{-0.6cm}
      \caption{\textbf{Illustration of knowledge-aware QA.} A sample question from CommonsenseQA can be better answered if a relevant subgraph of ConceptNet is provided as evidence. \textcolor{CornflowerBlue}{\textbf{Blue nodes}} correspond to entities mentioned in the question, \textcolor{LightPink}{\textbf{pink nodes}} correspond to those in the answer. The other nodes are some associated entities introduced when extracting the subgraph. $\star$ indicates the correct answer.}% For details, please refer to \sectionref{sec:formulation}. } 
      \label{fig:example}
%  \vspace{-0.2cm}
\end{figure}

Many recently proposed question answering  tasks require not only machine comprehension of the question and context, but also \textit{relational reasoning} over entities (concepts) and their relationships by referencing external knowledge~\cite{Talmor2018CommonsenseQAAQ, Sap2019SocialIQACR, clark2018think,Mihaylov2018CanAS}. 
For example, the question in Fig.~\ref{fig:example} requires a model to perform relational reasoning over mentioned entities, $\ie$, to infer latent relations among the concepts: $\{$\textsc{Child}, \textsc{Sit}, \textsc{Desk}, \textsc{Schoolroom}$\}$. 
Background knowledge such as \textit{``a child is likely to appear in a schoolroom''} may not be readily contained in the questions themselves, but are commonsensical to humans. 

Despite the success of large-scale pre-trained language models (PTLMs)~\cite{Devlin2019BERTPO,Liu2019RoBERTaAR}, 
% there is still a large performance gap between fine-tuned models and human performance on datasets that probe relational reasoning. 
these models fall short of providing interpretable predictions, as the knowledge in their pre-training corpus is not explicitly stated, but rather is implicitly learned.
It is thus difficult to recover the evidence used in the reasoning process.

This has led many to leverage knowledge graphs~(KGs)~\citep{Mihaylov2018KnowledgeableRE,kagnet-emnlp19,Wang2018ImprovingNL,yang-etal-2019-enhancing-pre}. 
KGs represent relational knowledge between entities with multi-relational edges for models to acquire.
Incorporating KGs brings the potential of interpretable and trustworthy predictions, as the knowledge is now explicitly stated. 
For example, in Fig.~\ref{fig:example}, the relational path $(\textsc{Child}\rightarrow \texttt{AtLocation}\rightarrow \textsc{Classroom}\rightarrow \texttt{Synonym}\rightarrow \textsc{Schoolroom})$ naturally provides evidence for the answer \textsc{Schoolroom}. 

\begin{figure}[t]
\vspace{-0.1cm}
      \centering
        \includegraphics[width=1.0\linewidth]{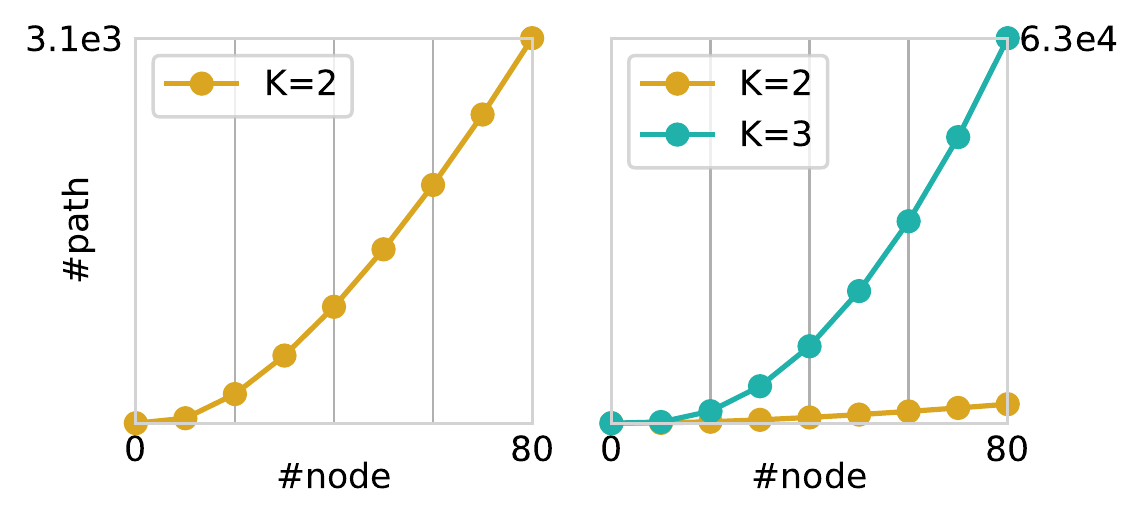}
    \vspace{-0.7cm}
      \caption{\textbf{Number of $K$-hop relational paths w.r.t. the node count in extracted graphs on  CommonsenseQA.} \textbf{Left}: The path count  is polynomial w.r.t. the number of nodes. \textbf{Right}: The path count is exponential w.r.t. the number of hops.}
      \label{fig:scalability}
\vspace{-0.0cm}
\end{figure}

A straightforward approach to leveraging a knowledge graph is to directly model these relational paths. KagNet~\citep{kagnet-emnlp19} and MHPGM~\citep{bauer-etal-2018-commonsense} model multi-hop relations by extracting relational paths from KG and encoding them with sequence models. 
Application of attention mechanisms upon these relational paths can further offer good interpretability. 
However, these models are hardly scalable because the number of possible paths in a graph is  (1) \textit{polynomial} w.r.t. the number of nodes (2) \textit{exponential} w.r.t. the path length (see Fig.~\ref{fig:scalability}). 
Therefore, some~\citep{weissenborn2017dynamic, Mihaylov2018KnowledgeableRE} resort to only using one-hop paths, namely, triples, to balance scalability and reasoning capacities.

Graph neural networks (GNNs), in contrast, enjoy better scalability via their message passing formulation, but usually lack transparency. 
The most commonly used GNNs' variant, Graph Convolutional Networks (GCNs)~\citep{Kipf2016SemiSupervisedCW}, perform message passing by aggregating neighborhood information for each node, but ignore the relation types.  RGCNs~\citep{Schlichtkrull2017ModelingRD} generalize GCNs by performing relation-specific aggregation, making it applicable to multi-relational graphs. 
However, these models do not distinguish the importance of different neighbors or relation types and thus cannot provide explicit relational paths for model behavior interpretation.

% 

% Alternatively, Relational GCN~\citep{Schlichtkrull2017ModelingRD} perform 
% Alternatively, Graph Neural Networks (GNNs) are designed to work directly on graph structures. Each node exchanges information with its neighboring nodes until an equilibrium is reached. Graph Convolutional Networks (GCNs)~\citep{Kipf2016SemiSupervisedCW} are variants of GNNs, where transformation matrices for message passing from neighbors are shared among all nodes. However, they can't model relational graphs as edge types are not taken into consideration. Relational Graph Convolutional Networks (RGCNs)~\citep{Schlichtkrull2017ModelingRD} use different transformation matrices for message passing on edges of different relational types, thus are capable of encoding multi-relational graphs. The message passing process for GNNs can be efficiently implemented as matrix multiplication, leading to good scalability. However, because messages are passed from direct neighbors and features are aggregated at node level, multi-hop logical paths, which are vital to reasoning, can hardly be interpreted from GNNs.

In this paper, we propose a novel graph encoding architecture, \textit{Multi-hop Graph Relation Network} (MHGRN), which
combines the strengths of path-based models and GNNs. 
Our model inherits scalability from GNNs by preserving the message passing formulation.
It also enjoys interpretability of path-based models by incorporating structured relational attention mechanism. 
Our key motivation is to perform \textit{multi-hop} message passing within a \textit{single} layer to allow each node to directly attend to its multi-hop neighbours, towards multi-hop relational reasoning.
We outline the favorable features of knowledge-aware QA models in Table~\ref{tab:property} and compare MHGRN with them.

\begin{table}[t]
\vspace{-0.0cm}
\small
\centering
\scalebox{0.85}{
\begin{tabular}{@{}cccccc@{}}
  \toprule
   &  {GCN}  & {RGCN} & KagNet & {MHGRN} \\ 
  \midrule
  Multi-Relational Encoding & \xmark  & \cmark & \cmark & \cmark \\
  Interpretable & \xmark & \xmark & \cmark & \cmark  \\
  Scalable w.r.t. \#node  & \cmark & \cmark  & \xmark & \cmark \\
  Scalable w.r.t. \#hop  & \cmark  & \cmark & \xmark & \cmark  \\
%   Multi-relational Reasoning  & \xmark & \xmark & \cmark & \cmark & \cmark \\
  \bottomrule
\end{tabular}
}
\vspace{-0.2cm}
\caption{\textbf{Properties} of our MHGRN and other representative models for graph encoding. }
\label{tab:property}
\vspace{-0.2cm}
\end{table}

We summarize the main contributions of this work as follows: 
1) We propose MHGRN, a novel model architecture tailored to multi-hop relational reasoning, which explicitly models multi-hop relational paths at scale.
2) We propose a structured relational attention mechanism for efficient and interpretable modeling of multi-hop reasoning paths, along with its training and inference algorithms.
3) We conduct extensive experiments on two question answering datasets and show that our models bring significant improvements compared to knowledge-agnostic PTLMs, and outperform other graph encoding methods by a large margin.

\section{Problem Formulation and Overview}
\label{sec:formulation}

\begin{figure}[t]
\vspace{-0.2cm}
      \centering
        \includegraphics[width=0.9\linewidth]{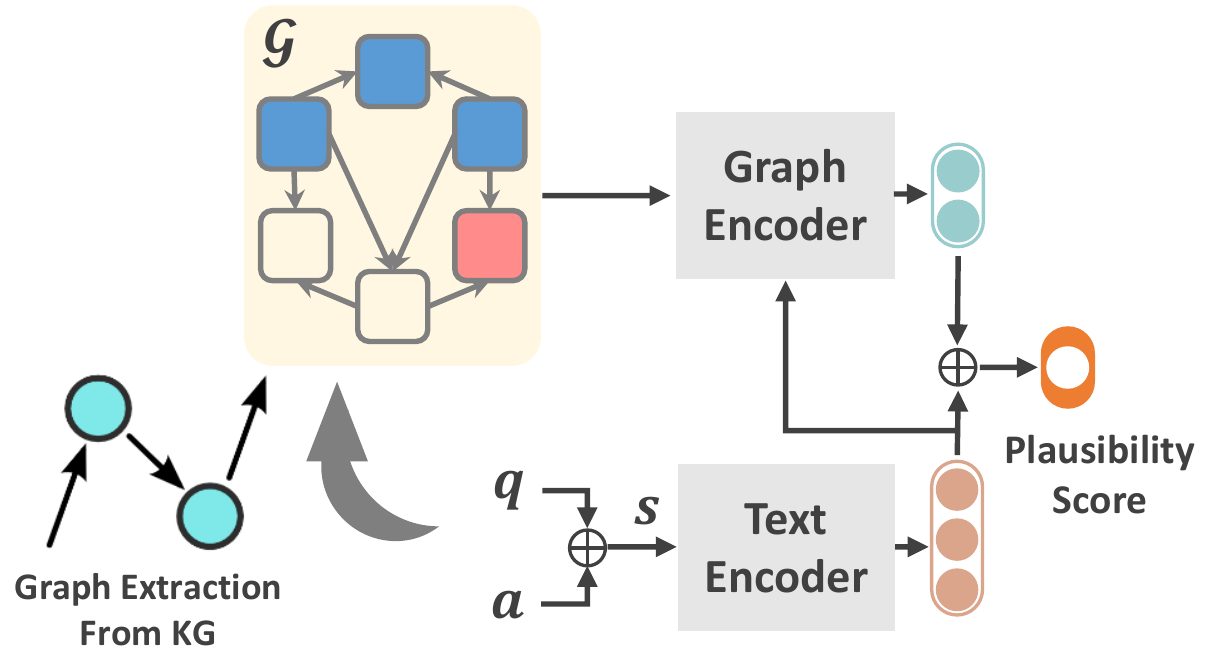}
        \vspace{-0.2cm}
      \caption{\textbf{Overview of the knowledge-aware QA framework.} It integrates the output from graph encoder (for relational reasoning over contextual subgraphs) and text encoder (for textual understanding) to generate the plausibility score for an answer option.}
      \label{fig:overview}
        \vspace{-0.2cm}
\end{figure}

In this paper, we limit the scope  to the task of multiple-choice question answering, although it can be easily generalized to other knowledge-guided tasks (\eg,~natural language inference). 
The overall paradigm of knowledge-aware QA is illustrated in Fig.~\ref{fig:overview}. 
Formally, given an external knowledge graph (KG) as the knowledge source and a question $q$, our goal is to identify the correct answer from a set $\mathcal{C}$ of given options. 
We turn this problem into measuring the plausibility score between $q$ and each option $a\in \mathcal{C}$ then selecting the one with the highest plausibility score.

Denote the vector representations of question $q$ and option $a$ as $\bm{q}$ and  $\bm{a}$.
To measure the score for $\bm{q}$ and $\bm{a}$, we first concatenate them to form a statement vector $\bm{s}=\left[\bm{q};\bm{a}\right]$. Then we extract from the external KG a subgraph $\mathcal{G}$ (\ie, schema graph in KagNet~\cite{kagnet-emnlp19}), with the guidance of $\bm{s}$ (detailed in \sectionref{subsec:exp:construct}). 
This contextualized subgraph is defined as a multi-relational graph $\mathcal{G} = (\mathcal{V}, \mathcal{E}, \phi)$.
Here $\mathcal{V}$ is a subset of entity in the external KG, containing only those relevant to $s$.  
$\mathcal{E} \subseteq \mathcal{V}\times \mathcal{R} \times \mathcal{V}$ is the set of edges that connect nodes in $\mathcal{V}$, where $\mathcal{R}=\{1, \cdots, m\}$ are ids of all pre-defined relation types. 
The mapping function $\phi(i): \mathcal{V} \rightarrow \mathcal{T}=\{\tE_{q}, \tE_{a}, \tE_{o}\}$ takes node $i\in V$ as input and outputs $\tE_{q}$ if $i$ is an entity mentioned in ${q}$, $\tE_{a}$ if it is mentioned in ${a}$, or $\tE_{o}$ otherwise. We finally encode the statement to $\bm{s}$, $\mathcal{G}$ to $\bm{g}$, concatenate $\bm{s}$ and $\bm{g}$, for calculating the plausibility score.

\section{Background: Multi-Relational Graph Encoding Methods}
\label{sec:background}

We leave encoding of $s$ to pre-trained language models and focus on the challenge of encoding graph $\mathcal{G}$ to capture latent relations between entities. Current methods for encoding multi-relational graphs mainly fall into two categories: GNNs and path-based models. GNNs encode structured information by passing messages between nodes, directly operating on the graph structure, while path-based methods first decompose the graph into paths and then pool features over them.

\smallskip
\noindent
\textbf{Graph Encoding with GNNs.}
For a graph with $n$ nodes, a graph neural network (GNN) takes a set of node features $\{\bm{h}_1, \bm{h}_2, \dots, \bm{h}_n\}$ as input, and computes their corresponding node embeddings $\{\bm{h}_1^\prime, \bm{h}_2^\prime, \dots, \bm{h}_n^\prime\}$ via message passing~\citep{gilmer2017neural}. A compact graph representation for $\mathcal{G}$ can thus be obtained by pooling over the node embeddings $\{\bm{h}_i^\prime\}$:
\begin{equation}
    \text{GNN}(\mathcal{G}) =  \text{Pool}(\{\bm{h}_1^\prime, \bm{h}_2^\prime, \dots, \bm{h}_n^\prime\}).
\end{equation}
As a notable variant of GNNs, graph convolutional networks (GCNs)~\citep{Kipf2016SemiSupervisedCW} additionally update node embeddings by aggregating messages from its direct neighbors. 
RGCNs~\citep{Schlichtkrull2017ModelingRD} extend GCNs to encode multi-relational graphs by defining relation-specific weight matrix $\bm{W}_{r}$ for each edge type:
\begin{equation} \label{eq:rgcn}
        \bm{h}_{i}^\prime=\sigma\left(\left(\sum_{r\in\mathcal{R}}|\mathcal{N}_i^r|\right)^{-1} \sum_{r\in\mathcal{R}} \sum_{j \in \mathcal{N}_{i}^{r}}  \bm{W}_{r}\bm{h}_j \right),
\end{equation}
where $\mathcal{N}_i^r$ denotes neighbors of node $i$ under relation $r$.\footnote{For simplicity, we assume a single graph convolutional layer. In practice, multiple layers are stacked to enable message passing from multi-hop neighbors.}

While GNNs have proved to have good scalability, their reasoning is done at the node level, making them incompatible with modeling paths. This property also hinders the model's decisions from being interpretable at the path level.

%Existing GNN generally adopts single-hop message passing mechanism \wj{Is this true? Do you mean on a QA task? You may give references.}, which hinders explicit multi-hop information flow, as a two-hop neighbor of node $i$ has no clue of the route from $i$ to itself and instead only knows about the representations of its direct neighbors.
% to-do

\smallskip
\noindent
\textbf{Graph Encoding with Path-Based Models.} In addition to directly modeling the graph with GNNs, a graph can also be viewed as a set of relational paths connecting pairs of entities.

Relation Networks (RNs)~\citep{Santoro2017ASN} can be adapted to multi-relational graph encoding under QA settings. RNs use MLPs to encode all triples (one-hop paths) in $\mathcal{G}$ whose head entity is in $\mathcal{Q}=\{j \mid \phi(j)=\tE_q\}$ and tail entity is in $\mathcal{A}=\{i \mid \phi(i)=\tE_a\}$. It then pools the triple embeddings to generate a vector for $\mathcal{G}$ as follows.
\begin{multline}
\label{eq:rn}
    \begin{small}\text{RN}(\mathcal{G}) =  \text{Pool}\Big(\{\text{MLP}(\bm{h}_j\oplus \bm{e}_r \oplus\end{small} \\
     \begin{small}\bm{h}_i ) \mid 
   j\in \mathcal{Q}, i\in \mathcal{A}, (j, r, i)\in \mathcal{E}\}\Big)\end{small}.
\end{multline}
Here $\bm{h}_j$ and $\bm{h}_i$ are features for nodes $j$ and $i$, $\bm{e}_r$ is the embedding of relation $r\in\mathcal{R}$, $\oplus$ denotes vector concatenation.
%and $\mathcal{Q}$ and $\mathcal{A}$ are chosen as $\mathcal{Q}=\{j \mid \phi(j)=\tE_q\}$ and $\mathcal{A}=\{i \mid \phi(i)=\tE_a\}$ respectively\footnote{For general NLP tasks, simply choose both sets as the set of mentioned entities.}.\jun{need to talk about generalizing (3) to multi-hop r / path}

% \wj{RN only encodes one-hop path? I can't find information for multihop path here. If RN only encodes one-hop path, then can we say it is a path-based method? I think it should cover multi-hop paths.} 

% RN works explicitly with paths and can be easily augmented with attention mechanism to identify important reasoning chains, thus has good interpretability. However, the computation cost grows exponentially with respect to path length, limiting its scalability.

To further equip RN with the ability to model nondegenerate paths, KagNet~\citep{kagnet-emnlp19} adopts LSTMs to encode all paths connecting question entities and answer entities with lengths no more than $K$. It then aggregates all path embeddings via attention mechanism:
\begin{multline}
\label{eq:kagnet}
    \begin{small}\textsc{KagNet}(\mathcal{G}) =  \text{Pool}\Big(\{\text{LSTM}(j, r_1, \dots, r_k, i) \mid \end{small} \\
     \begin{small}(j, r_1, j_1), \cdots, (j_{k-1}, r_k, i) \in \mathcal{E}, 1 \leq k\leq K\}\Big) \end{small}.
\end{multline}

%is an extension of RN. It extracts paths with lengths less than $k$ that connect a question entity and an answer entity in $\mathcal{G}$, and encodes the paths using LSTMs, after which the path representations are aggregated via attention mechanism.

% \wj{As far as I know, RN part in KagNet does not encode ``path". It just encodes a concept pair.}

%Although path-based methods can explicitly model reasoning paths between entities, they cannot scale as path length grows, as the computation cost grows exponentially with respect to $k$. As a consequence, KagNet relies on extensive path pruning to make computation feasible, which on the other hand may filter out useful information in the graph.
%\wj{I think you don't have to mention KagNet here unless KagNet is a base model of your method. You may describe KagNet in Relate work.}
%%%%%%%%%%%%%%%%%%%%%%%%%%%%%%%%%
%          Methodology          %
%%%%%%%%%%%%%%%%%%%%%%%%%%%%%%%%%
\section{Proposed Method: Multi-Hop Graph Relation Network (MHGRN)}
\label{sec:method}
% To summarize existing graph encoding approaches, path based models are interpretable but not scalable with respect to path length, while GNN based models lack interpretablity and do not reach competitive performance. 
This section presents \textit{Multi-hop Graph Relation Network} (MHGRN), a novel GNN architecture that unifies both GNNs and path-based models. MHGRN inherits path-level reasoning and interpretabilty from path-based models, while preserving good scalability of GNNs. 
% which is tailored for modeling multi-relational graphs. Our model is scalable, interpretable and achieve competitive performance. It follows the message passing framework like previous GNNs, but is endowed with larger model capacity by incorporating: (1) node type specific transformation (2) structured attention mechanism designed for relational data. With part of its parameters frozen, our model recovers relation network or relation GCN as special cases. 
% path reasoning? your defined RN is one-hop

\begin{figure}[t]
\vspace{-0.0cm}
      \centering
        \includegraphics[width=0.96\linewidth]{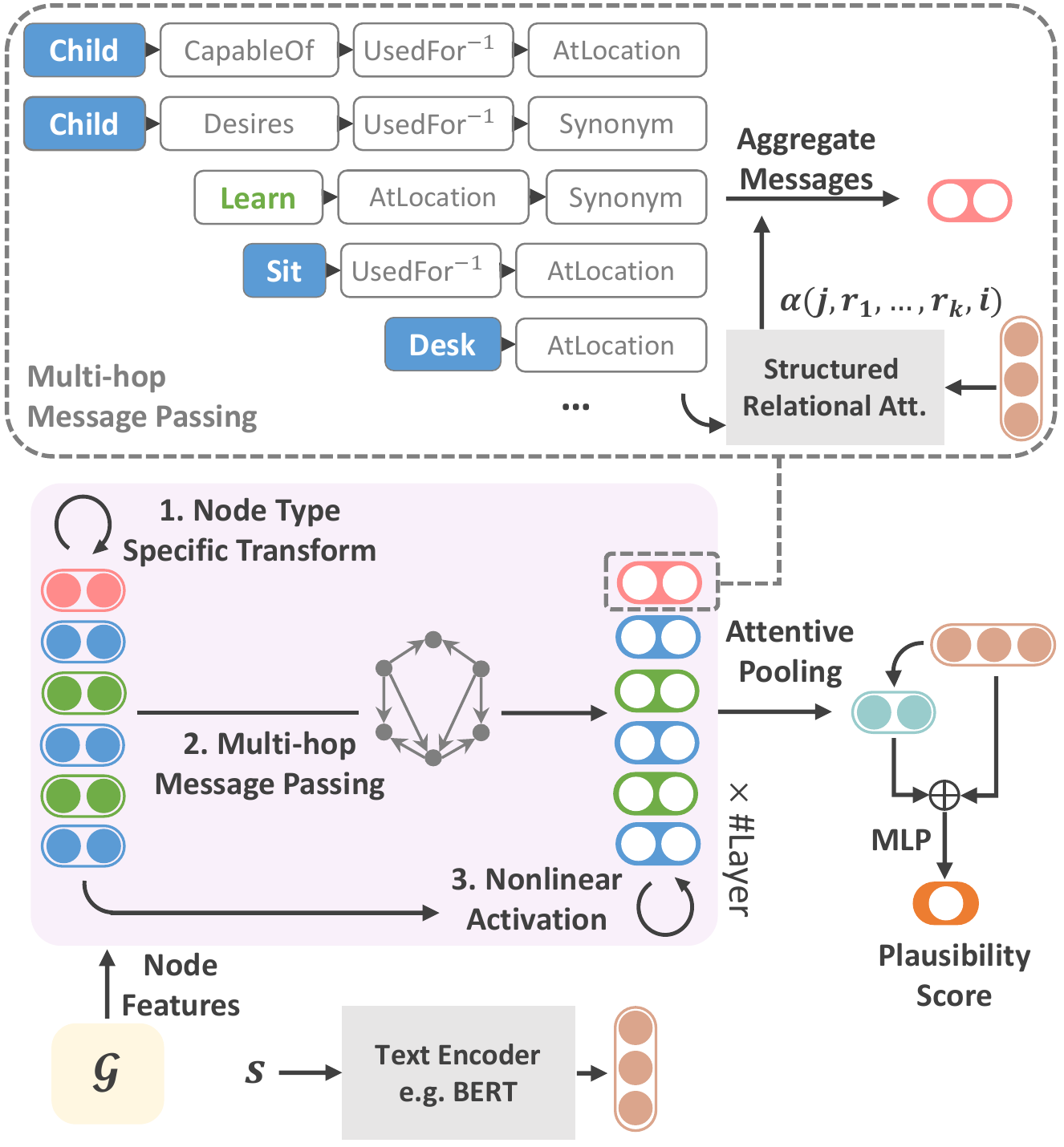}
        \vspace{-0.0cm}
      \caption{\textbf{Our proposed MHGRN architecture for relational reasoning.} MHGRN takes a multi-relational graph $\mathcal{G}$ and a (question-answer) statement vector $\bm{s}$ as input, and outputs a scalar that represent the plausibility score of this statement.}
      \label{fig:model}
     \vspace{-0.0cm}
\end{figure}

% In the rest of this section, we provide a detailed introduction of MHGRN. We first present its overall architecture (\sectionref{subsec:method:arch}) and elaborate its structured attention component (\sectionref{subsec:method:att}). We then analyze its computation complexity (\sectionref{subsec:method:complexity}) and its expressive power (\sectionref{subsec:method:power}), and finally discuss the training and inference algorithms for multiple-choice question answering (\sectionref{subsec:method:learning}).
% In the rest of the section, we will provide detailed introduction of MHGRN architecture, including its capability of incorporating (1) node type specific transformation ; and (2) structured attention mechanism designed for multi-relational data, and discuss why such model can yield more expressive power as compared to GNNs and RNs.

\subsection{MHGRN: Model Architecture} \label{subsec:method:arch}
We follow the GNN framework introduced in \sectionref{sec:background}, where node features can be initialized with pre-trained weights (details in Appendix~\ref{app:details}). Here we focus on the computation of node embeddings.

\smallskip
\noindent
\textbf{Type-Specific Transformation.} 
To make our model aware of the node type $\phi$, we first perform node type specific linear transformation on the input node features:
\begin{equation} \label{eq:model_type_transorm}
\bm{x}_i = \bm{U}_{\phi(i)} \bm{h}_i + \bm{b}_{\phi(i)},
\end{equation}
where the learnable parameters $\bm{U}$ and $\bm{b}$ are specific to the type of node $i$. %This can be regarded as a more general way to incorporate node type information than directly adding type embeddings to the input. If the $\bm{U}$ matrices are set to identity matrices then it reduces to adding type embedding to the input node features. By performing non-identity transformations our model learns to distinguish the importance of different node types (e.g. it may learn to reduce the importance of non-mentioned concepts by learning a low rank $\bm{U}_{\tE_{o}}$).

\smallskip
\noindent
%In contrast to conventional GNNs where multi-hop message passing is enabled by stacking multiple single-hop message passing layers thus making the paths implicit
\textbf{Multi-Hop Message Passing.}
% \wj{I think this is a big contribution of this paper. But I feel there is no enough stress on this. You may elaborate this and stress more. By the way, what is difference between multi-layered RGCN and this one? I think this is RGCN+RN? RG}
As mentioned before, our motivation is to endow GNNs with the capability of \textit{directly modeling paths}. To this end, we propose to pass messages directly over all the relational paths of lengths up to $K$. The set of valid $k$-hop relational paths is defined as:
\begin{multline}
    \begin{small}\Phi_k = \{(j, r_1, \dots, r_k, i) \mid (j, r_1, j_1),\end{small} \\ 
    \begin{small}\cdots, (j_{k-1}, r_k, i) \in \mathcal{E}\} \quad (1\le k\le K)\end{small}.
\end{multline}
We perform $k$-hop \begin{small}($1\le k\le K$)\end{small} message passing over these paths, which is a generalization of the single-hop message passing in RGCNs (see Eq.~\ref{eq:rgcn}):
% \wj{I think this is not a simple extension of RGCN. This is RGCN+RN. You may need to stress this and remove "extension of single-hop RGCN". This word is  confusing and makes this look incremental work.}
\begin{multline}
        \label{eq:model_message_pass}
        \begin{small}\bm{z}_i^k = \sum_{(j, r_1,\dots,r_k, i)\in \Phi_k} \alpha(j, r_1, \dots, r_k, i)/d_i^k ~\cdot~ \bm{W}_{0}^{K}\end{small} \\
        \begin{small}\cdots\bm{W}_{0}^{k+1}  \bm{W}_{r_k}^k\cdots\bm{W}_{r_1}^1 \bm{x}_j \quad (1\le k\le K)\end{small},
\end{multline}
where the \begin{small}$\bm{W}^t_r (1\le t \le K, 0\le r \le m)$\end{small} matrices are learnable\footnote{$\bm{W}_0^t (0\le t \le K)$ are introduced as padding matrices so that $K$ transformations are applied regardless of $k$, thus ensuring comparable scale of $\bm{z}_i^k$ across different $k$.}, $\alpha(j, r_1, \dots, r_k, i)$ is an attention score elaborated in \sectionref{subsec:method:att} and \begin{small}$d_i^k=\sum_{(j\cdots i)\in \Phi_k} \alpha(j\cdots i)$\end{small} is the normalization factor. The \begin{small}$\{\bm{W}_{r_k}^k \cdots \bm{W}_{r_1}^1 \mid 1\le r_1,\dots,r_k \le m\}$\end{small} matrices can be interpreted as the low rank approximation of a \{$m\times \cdots \times m\}_{k}\times d\times d$ tensor that assigns a separate transformation for each $k$-hop relation, where $d$ is the dim. of $\bm{x}_i$. 

Incoming messages from paths of different lengths are aggregated via attention mechanism~\citep{vaswani2017attention}: %\wj{what is $s$? It's hard to find the definition. Also, what is bilinear? You should define every terms to make the paper self-contained.}
\begin{equation} \label{eq:bilinear}
    \bm{z}_i = \sum_{k=1}^K \text{softmax}\big(\text{bilinear}\big(\bm{s}, \bm{z}_i^k\big)\big) \cdot \bm{z}_i^k.
\end{equation}
% \smallskip    
\noindent
\textbf{Non-linear Activation.} Finally, we apply shortcut connection %\wj{between where?} 
and nonlinear activation to obtain the output node embeddings.%\xiang{elaborate a bit on why such design; can add refs if needed.}
\begin{equation}
        \label{eq:model_nonlinear}
        \bm{h}_i^\prime=\sigma\left(\bm{V}\bm{h}_i  + \bm{V}^\prime\bm{z}_i \right),
\end{equation}
where $\bm{V}$ and $\bm{V}^\prime$ are learnable model parameters, and $\sigma$ is a non-linear activation function.

\subsection{Structured Relational Attention} \label{subsec:method:att}
Here we work towards effectively parameterizing the attention score \begin{small}$\alpha(j, r_1,\dots, r_k, i)$ in Eq.~\ref{eq:model_message_pass}\end{small} for all $k$-hop paths without introducing \begin{small}$\mathcal{O}(m^k)$\end{small} parameters. We first regard it as the probability of a relation sequence \begin{small}$\left( \phi(j), r_1, \dots, r_k, \phi(i) \right)$\end{small} conditioned on $\bm{s}$:
\begin{equation} \label{eq:att_prob}
        \alpha(j, r_1,\dots, r_k, i) = p\left( \phi(j), r_1, \dots, r_k, \phi(i) \mid \bm{s}\right),
    \end{equation}
which can naturally be modeled by a probabilistic graphical model, such as conditional random field \citep{Lafferty2001ConditionalRF}:
\begin{multline} \label{eq:att_crf}
    p\left( \cdots \mid \bm{s}\right) \propto \exp\Bigg(f(\phi(j), \bm{s})+ \sum_{t=1}^k \delta({r_t}, \bm{s}) \\
    + \sum_{t=1}^{k-1} \tau({r_t}, {r_{t+1}}) + g(\phi(i), \bm{s}) \Bigg) \\
    \overset{\Delta}{=}  \underbrace{\beta(r_1, \dots, r_k, \bm{s})}_{\text{Relation Type Attention}} \cdot  \underbrace{\gamma(\phi(j),\phi(i), \bm{s})}_{\text{Node Type Attention}},
\end{multline}
where $f(\cdot)$, $\delta(\cdot)$ and $g(\cdot)$ are parameterized by two-layer MLPs and $\tau(\cdot)$ by a transition matrix of shape $m\times m$. Intuitively, $\beta(\cdot)$ models the importance of a $k$-hop relation while $\gamma(\cdot)$ models the importance of messages from node type $\phi(j)$ to $\phi(i)$ (\eg,~the model can learn to pass messages only from question entities to answer entities).

Our model scores a $k$-hop relation by decomposing it into both context-aware single-hop relations (modeled by $\delta$) and two-hop relations (modeled by $\tau$). We argue that $\tau$ is indispensable,  without which the model may assign high importance to illogical multi-hop relations (\eg,~[\texttt{AtLocation}, \texttt{CapableOf}]) or  noisy relations (\eg,~[\texttt{RelatedTo}, \texttt{RelatedTo}]).

% For example, even though the relation \texttt{CapableOf} may be highly relevant to $\bm{s}$, the illogical two-hop relation $(\texttt{CapableOf}, \texttt{CapableOf})$ should still be assigned a low attention score.

% The rationale for introducing transition scores $\tau(r_i, r_{i+1})$ is that the model cannot distinguish certain $k$-hop relations using only the unary scores. For example, even though the relation \texttt{CapableOf} may be relevant to $\bm{s}$, the two-hop relation $(\texttt{CapableOf}, \texttt{CapableOf})$ should still be assigned a low attention score since it is logically incorrect.

% While the number of parameters introduced is only $\mathcal{O}(m^2)$ which is independent of $k$, a $k$-hop relation can be approximated by the composition of two-hop relations (modeled by $\tau$) thus making its structure locally preserved.

% to-do

% The rationale for introducing transition scores $\tau(r_i, r_{i+1})$ is that the model cannot distinguish the importance of certain $k$-hop relations with only the unary scores from $\delta$. For example, even though the relation $\text{CapableOf}$ may be relevant to the sentence, the two-hop relation $(\text{CapableOf}, \text{CapableOf})$ should be assigned a low attention score since it is semantically incorrect.

\subsection{Computation Complexity Analysis} \label{subsec:method:complexity}
% Although our model needs to aggregate messages from potentially exponential number of paths in Eq.~\ref{eq:model_message_pass}
Although message passing process in Eq.~\ref{eq:model_message_pass} and attention module in Eq.\ref{eq:att_crf} handles potentially exponential number of paths, it can be computed in linear time with dynamic programming (see Appendix~\ref{app:dp}). As summarized in Table~\ref{tab:complexity}, time complexity and space complexity of MHGRN on a sparse graph are  both linear w.r.t. either the maximum path length $K$ or the number of nodes $n$. 
% In contrast, the complexity of path-based models (\eg~KagNet) is exponential with respect to $K$ and polynomial with respect to $n$, making them hardly scalable as the path length or the graph size increases.

% \qinyuan{One thing that reviewers may challenge is whether your scalability is tested empirically in the experiments. In fig 1 the largest subgraph has $n=80$ nodes? if i'm understanding it correctly. What happens if $n$ or $K$ is even larger?}

% \begin{center}
\begin{table}[t]
\vspace{-0.0cm}
    \centering
     \scalebox{0.82}{
     \begin{tabular}{lcc}
      \toprule
      \textbf{Model} & \textbf{Time}  & \textbf{Space}  \\
      \midrule
      \multicolumn{3}{c}{\textit{$\mathcal{G}$ is a dense graph}} \\
      \midrule
      $K$-hop KagNet & $\mathcal{O}\left(m^K n^{K+1}K \right)$ & $\mathcal{O}\left(m^K n^{K+1}K \right)$ \\
      $K$-layer RGCN & $\mathcal{O}\left ( mn^2K \right)$ & $\mathcal{O}\left( mnK \right)$ \\
      MHGRN & $\mathcal{O}\left(m^2n^2K \right)$ &  $\mathcal{O}\left ( mnK \right)$ \\
      \midrule
      \multicolumn{3}{c}{\textit{$\mathcal{G}$ is a sparse graph with maximum node degree $\Delta \ll n$}} \\
      \midrule
      $K$-hop KagNet & $\mathcal{O}\left(m^KnK\Delta^{K} \right)$ &  $\mathcal{O}\left(m^KnK\Delta^{K} \right)$ \\
       $K$-layer RGCN & $\mathcal{O}\left(mnK\Delta \right)$ & $\mathcal{O}\left ( mnK \right)$ \\
      MHGRN & $\mathcal{O}\left(m^2n K\Delta \right)$ &  $\mathcal{O}\left ( mnK \right)$ \\
    \bottomrule
    \end{tabular}
    }
    \vspace{-0.0cm}
    \caption{\textbf{Computation complexity} of different $K$-hop reasoning models on a dense/sparse multi-relational graph with $n$ nodes and $m$ relation types. Despite the quadratic complexity w.r.t. $m$, MHGRN's time cost is similar to RGCN on GPUs with parallelizable matrix multiplications (cf. Fig.~\ref{fig:scale}).} \label{tab:complexity}
\end{table}
% \end{center}

    % \footnote{To gain some intuition why it can be computed in linear time. We can consider the simplified version where the $\beta(\cdots)$, $\bm{g}$ and $\bm{g}^\prime$ are removed. In this case, Eq. \ref{eq:model_matrix} reduces to a $k$-layer relational GCN without nonlinearity (see Eq. \ref{eq:rgcn}), in which case the computation complexity is $\mathcal{O}(pmk)$, where $p$ is the maximum edge number for each relation type.} 
% \input{algorithms/message_passing.tex}

\begin{table*}[t]
\centering
\scalebox{0.71}{
\begin{tabular}{lcccccc}
    \toprule  
    \multirow{2}{*}{\textbf{Methods}}&
    \multicolumn{2}{c}{ \textbf{BERT-Base}}&\multicolumn{2}{c}{ \textbf{BERT-Large}}&\multicolumn{2}{c}{ \textbf{RoBERTa-Large}}\\
    \cmidrule(lr){2-3} \cmidrule(lr){4-5} \cmidrule(lr){6-7}
     & IHdev-Acc.(\%) & IHtest-Acc.(\%) & IHdev-Acc.(\%) & IHtest-Acc.(\%) & IHdev-Acc.(\%) & IHtest-Acc.(\%) \\
    \midrule  
    {w/o KG}  &  57.31~($\pm$1.07)  &  53.47~($\pm$0.87)  &  61.06~($\pm$0.85) & 55.39~($\pm$0.40) & 73.07~($\pm$0.45) & 68.69($\pm$0.56) \\
    \midrule
    {RGCN~\cite{Schlichtkrull2017ModelingRD}} &  56.94~($\pm$0.38)  &  54.50~($\pm$0.56)  &  62.98~($\pm$0.82) & 57.13~($\pm$0.36) & 72.69~($\pm$0.19) & 68.41~($\pm$0.66) \\
    {GconAttn~\citep{Wang2018ImprovingNL}}  & 57.27~($\pm$0.70) & 54.84~($\pm$0.88) & 63.17~($\pm$0.18) &  57.36~($\pm$0.90) & 72.61($~\pm$0.39) &
    68.59~($\pm$0.96)\\
    % {KVM~\citep{Mihaylov2018KnowledgeableRE}}  & 57.19~($\pm$0.68) & 52.88~($\pm$0.52) & 62.51($\pm$0.62) &  57.42~($\pm$0.33)  & 74.53($~\pm$0.29)   & 70.37~($\pm$0.69)   \\
    {KagNet$^{\dagger}$~\citep{kagnet-emnlp19}}  &  55.57  &  56.19  &  62.35 & 57.16 & - & - \\
    {RN (1-hop)}  &  58.27~($\pm$0.22)  &  56.20~($\pm$0.45)  &  63.04~($\pm$0.58) & 58.46~($\pm$0.71) & 74.57~($\pm$0.91) & 69.08~($\pm$0.21) \\
    {RN (2-hop)} &  59.81~($\pm$0.76)  & 56.61~($\pm$0.68)  &  63.36~($\pm$0.26) & 58.92~($\pm$0.14) & 73.65~($\pm$3.09) & 69.59~($\pm$3.80) \\
    \midrule
    % {MHGRN ($K=1$)} & 60.40(~$\pm$0.64) & 56.83~($\pm$0.54) & 63.91($\pm$0.42) &  60.33~($\pm$0.69)  & 73.24~($\pm$0.55)   & 70.45~($\pm$1.04)  \\
    % {MHGRN ($K=2$)} & 60.36~($\pm$0.23) & \textbf{57.23}~($\pm$0.82) & 63.29($\pm$0.51) &  \textbf{60.59}~($\pm$0.58)  & 73.93~($\pm$0.77)   & 70.83~($\pm$0.50)  \\
    {MHGRN} & 60.36~($\pm$0.23) & \textbf{57.23}~($\pm$0.82) & 63.29($\pm$0.51) &  \textbf{60.59}~($\pm$0.58)  & 74.45~($\pm$0.10)   & \textbf{71.11}~($\pm$0.81)  \\
    \bottomrule 
\end{tabular}
}
\vspace{-0.1cm}
\caption{\textbf{Performance comparison on CommonsenseQA in-house split.} We report in-house Dev (IHdev) and Test (IHtest) accuracy (mean and standard deviation of four runs) using the data split of \citet{kagnet-emnlp19} on CommonsenseQA. $\dagger$ indicates reported results in its paper.}
\label{tab:results:csqa}
\end{table*}

\subsection{Expressive Power of MHGRN} \label{subsec:method:power}
In addition to efficiency and scalability, we now discuss the modeling capacity of MHGRN. With the message passing formulation and relation-specific transformations, it is by nature the generalization of RGCN. It is also capable of directly modeling paths, making it interpretable as are path-based models like RN and KagNet. To show this, we first generalize RN (Eq.~\ref{eq:rn}) to the multi-hop setting and introduce $K$-hop RN (formal definition in Appendix~\ref{app:rn}), which models multi-hop relation as the composition of single-hop relations. We show that MHGRN is capable of representing $K$-hop RN (proof in Appendix \ref{app:proof}).

%Next we can establish the following theorem to show that MHGRN is capable of representing $K$-hop RN (see Appendix \ref{app:proof} for a proof).

\subsection{Learning, Inference and Path Decoding}
\label{subsec:method:learning}

We now discuss the learning and inference process of MHGRN instantiated for QA tasks. Following the problem formulation in ~\sectionref{sec:formulation}, we aim to determine the plausibility of an answer option $\bm{a}\in \mathcal{C}$ given the question $\bm{q}$ with the information from both text $s$ and graph $\mathcal{G}$. We first obtain the graph representation $\bm{g}$ by performing attentive pooling over the output node embeddings of answer entities $\{\bm{h}_i^\prime\mid i\in \mathcal{A}\}$. Next we concatenate it with the text representation $\bm{s}$ and compute the plausibility score by $\rho(\bm{q}, \bm{a}) = \texttt{MLP}(\bm{s}\oplus \bm{g})$.

During training, we maximize the plausibility score of the correct answer $\hat{\bm{a}}$ by minimizing the \textit{cross-entropy} loss:
\begin{equation}
    \mathcal{L} = \mathbb{E}_{\bm{q}, \hat{\bm{a}}, \mathcal{C}}\left[-\log\frac{\exp(\rho(\bm{q}, \hat{\bm{a}}))}{\sum_{\bm{a} \in \mathcal{C}}\exp(\rho(\bm{q}, \bm{a}))}\right].
\end{equation}
The whole model is trained \textbf{end-to-end} jointly with the text encoder (e.g., RoBERTa).

During inference, we predict the most plausible answer by $\text{argmax}_{\bm{a}\in \mathcal{C}}~\rho(\bm{q}, \bm{a})$. Additionally, we can decode a reasoning path as evidence for model predictions, endowing our model with the interpretability enjoyed by path-based models. Specifically, we first determine the answer entity $i^{*}$ with the highest score in the pooling layer and the path length $k^{*}$ with the highest score in Eq.~\ref{eq:bilinear}. Then the reasoning path is decoded by $\text{argmax}~\alpha(j, r_1, \dots, r_{k^{*}}, i^{*})$, which can be computed in linear time using dynamic programming.

% \xiang{1. This section is too rough now, need to enhance. Use a symbol to replace ``Pls". 2. Add footnote to clarify the input and parameters of MHGRN. 3. Need to update Eq (11) to adapt to QA case and clarify the candidate set is per question.}

% We This section presents the model architecture and loss function instantiated for the task of natural language inference.

% We compute plausibility score of a premise $\bm{p}$ and a hypothesis $\bm{h}$ by:
% \begin{equation}
%     Pls(\bm{p}, \bm{h}) = \textit{MLP}(\bm{s}\oplus\textit{MHGRN}(\mathcal{G}, \bm{s})),
% \end{equation}
% where $\bm{s}$ is the vector representation of the concatenation of $\bm{p}$ and $\bm{h}$ obtained from a text encoder and $\mathcal{G}$ is an extracted multi-relational graph conditioned on $\bm{p}$ and $\bm{h}$. During training, the model is presented with a premise $\bm{p}$, and a set of hypothesis $\mathcal{C} = \{\bm{h}_1, \bm{h}_2, \dots\}$ among which $\hat{\bm{h}}$ is the most plausible one. The parameters are learned by minimizing the cross-entropy loss:
% \begin{equation}
%     \mathcal{L} = \mathbb{E}\left[\frac{\exp(\zeta(\bm{p}, \hat{\bm{h}}, ))}{\sum_{\bm{h}^\prime \in \mathcal{HS}}\exp(Pls(\bm{p}, \bm{h}^\prime, ))}\right]
% \end{equation}

% During inference, the most plausible hypothesis given a premise $\bm{p}$ is selected by:
% \begin{equation}
%     \text{argmax}_{\bm{h}\in \mathcal{HS}}Pls(\bm{p}, \bm{h})
% \end{equation}
\section{Experimental Setup}
\label{sec:exp}
\vspace{-0.1cm}
We introduce how we construct $\mathcal{G}$ (\sectionref{subsec:exp:construct}), the datasets (\sectionref{subsec:exp:datasets}), as well as the baseline methods (\sectionref{subsec:exp:baselines}). 
Appendix~\ref{app:details} shows more implementation and experimental details for reproducibility.

\vspace{-0.1cm}
\subsection{Extracting $\mathcal{G}$ from External KG}
\label{subsec:exp:construct}
We use \textit{ConceptNet}~\citep{Speer2016ConceptNet5A}, a general-domain knowledge graph as our external KG to test models' ability to harness structured knowledge source. Following KagNet~\cite{kagnet-emnlp19}, we merge relation types to increase graph density and add reverse relations to construct a multi-relational graph with 34 relation types~(details in Appendix \ref{app:merge}).
% To extract a informative graph $\mathcal{G}$ from the KG, we first identify the entity mentions in $\bm{s}$ to entities in \textit{ConceptNet} to produce an initial entity set $\mathcal{V}$. We then add to $\mathcal{V}$ all the entities that appear in any 2-hop paths between pairs of mentioned entities and reserve all the edges between nodes, forming a subgraph $\mathcal{G}$.
To extract an informative contextualized graph $\mathcal{G}$ from KG, we recognize entity mentions in $s$ and link them to entities in \textit{ConceptNet}, with which we initialize our node set $\mathcal{V}$. We then add to $\mathcal{V}$ all the entities that appear in any two-hop paths between pairs of mentioned entities. Unlike KagNet, we do not perform any pruning but instead reserve all the edges between nodes in $\mathcal{V}$, forming our $\mathcal{G}$.

\subsection{Datasets}
\label{subsec:exp:datasets}
We evaluate models on two multiple-choice question answering datasets, CommonsenseQA and OpenbookQA.
Both require world knowledge beyond textual understanding to perform well.

% \begin{table}
% \vspace{-0.2cm}
% \centering
% \scalebox{0.68}{
% \begin{tabular}{lcc}
%     \toprule
%     \textbf{Methods} & \textbf{BERT-Base} & \textbf{BERT-Large} \\
%     \midrule
%     w/o KG & 53.57 & 62.34  \\
%     AMS~\textsuperscript{\dag}\citep{ye2019align} & - & 59.1$^\dagger$ \\
%     CS-Pretrain\textsuperscript{\dag}~\citep{li2019teaching} & 59.28 ($\pm$0.43)$^\dagger$ & - \\
%     \midrule
%     GconAttn & 58.50~($\pm$0.82) & 63.63($\pm$0.41) \\
%     KVM & 58.21~($\pm$0.35) & 62.68~($\pm$0.39) \\
%     % RN (1-hop, attn pool) & xx.xx~($\pm$0.xx) & xx.xx~($\pm$0.xx)  \\
%     RGCN & 57.84~($\pm$0.55) & 63.49~($\pm$0.31) \\
%     \midrule
%     MHGRN ($K=2$)& \textbf{59.87}~($\pm$0.53) & \textbf{64.56}~($\pm$0.18) \\
%     \bottomrule
% \end{tabular}
% }
% \vspace{-0.2cm}
% \caption{Performance comparison (accuracy in \%) on official Dev of CommonsenseQA with baselines (reported on leaderboard). $\dagger$ indicates reported results.}
% \label{tab:results:csqa_official}
% \vspace{-0.2cm}
% \end{table}
\begin{table}
% \vspace{-0.2cm}
\centering
\scalebox{0.68}{
\begin{tabular}{@{}lcc@{}}
    \toprule
    \textbf{Methods} & \textbf{Single} & \textbf{Ensemble} \\
    \midrule
    UnifiedQA\textsuperscript{\dag}~\cite{khashabi2020unifiedqa} & \textbf{79.1} & -\\
    \midrule
    RoBERTa\textsuperscript{\dag} & 72.1 & 72.5  \\
    RoBERTa + KEDGN\textsuperscript{\dag} & 72.5 & 74.4 \\
    RoBERTa + KE\textsuperscript{\dag} & 73.3 & - \\
    RoBERTa + HyKAS 2.0\textsuperscript{\dag}~\citep{Ma_2019} & 73.2 & - \\
    RoBERTa + FreeLB\textsuperscript{\dag}\cite{Zhu2020FreeLB} & 72.2 & 73.1\\
    XLNet + DREAM\textsuperscript{\dag}~ &	66.9 &	73.3 \\ 
    XLNet + GR\textsuperscript{\dag}~\citep{lv2019graphbased} & 75.3 & - \\
    % RN (1-hop, attn pool) & xx.xx~($\pm$0.xx) & xx.xx~($\pm$0.xx)  \\
    ALBERT\textsuperscript{\dag}~\cite{Lan2019ALBERTAL} & - & {\textbf{76.5}} \\
    % AristoBERTv7 \textsuperscript{\dag}~ & 64.6 & -\\
    \midrule
    RoBERTa + MHGRN ($K=2$)& 75.4 & \textbf{76.5} \\
    \bottomrule
\end{tabular}
}
% \vspace{-0.2cm}
\caption{\textbf{Performance comparison on official test of CommonsenseQA} with leaderboard SoTAs\footnote{Models based on ConceptNet are no longer shown on the leaderboard, and we got our results from the organizers.} (accuracy in \%). $\dagger$ indicates reported results on leaderboard. UnifiedQA uses T5-11B as text encoder, whose number of parameters is about 30 times more than other models.}
\label{tab:results:csqa_official}
\vspace{-0.2cm}
\end{table}

\textbf{CommonsenseQA}~\citep{Talmor2018CommonsenseQAAQ} necessitates various commonsense reasoning skills. 
The questions are created with entities from \textit{ConceptNet} and they are designed to probe latent compositional relations between entities in \textit{ConceptNet}.

% For this dataset, we use the in-house data split from \cite{kagnet-emnlp19}. 
\textbf{OpenBookQA}~\citep{Mihaylov2018CanAS} provides %$5,957$ $(4,957/500/500)$
elementary science questions together with an open book of science facts. 
This dataset also probes general common sense beyond the provided facts. 
% As labels of test examples are readily provided, we use the official split for evaluation.

\subsection{Compared Methods}
\label{subsec:exp:baselines}
%We compare our model with both knowledge-agnostic fine-tuning of pre-trained LMs and models that incorporate KG as external sources. 
We implement both knowledge-agnostic fine-tuning of pre-trained LMs and models that incorporate KG as external sources as our baselines. Additionally, we directly compare our model with the results from corresponding leaderboard. 
These methods typically leverage textual knowledge or extra training data, as opposed to external KG.
In all our \textit{implemented} models, 
we use pre-trained LMs as text encoders for $s$ for fair comparison. We do compare our models with those~\citep{Ma_2019,lv2019graphbased,khashabi2020unifiedqa} augmented by other text-form external knowledge (\eg, Wikipedia), although we stick to our focus of encoding \textit{structured} KG. 

Specifically, we fine-tune {\textsc{Bert-Base}}, {\textsc{Bert-Large}}~\citep{Devlin2019BERTPO}, and {\textsc{RoBERTa}}~\citep{Liu2019RoBERTaAR} for multiple-choice questions. We take \texttt{RGCN} (Eq.~\ref{eq:rgcn} in \sectionref {sec:background}), \texttt{RN}\footnote{We use mean pooling for 1-hop RN and attentive pooling for 2-hop RN (detailed in Appendix~\ref{app:rn}).} (Eq.~\ref{eq:rn} in ~\sectionref{sec:background}), \texttt{KagNet} (Eq.~\ref{eq:kagnet} in ~\sectionref{sec:background}) and \texttt{GconAttn}~\citep{Wang2018ImprovingNL} as baselines.
% We argue that these models, either fine-tuning pre-trained LMs with knowledge-rich text or datasets before fine-tuning them on the target dataset, or fine-tuning with additional text as evidence, are all orthogonal to our work.
\texttt{GconAttn} generalizes match-LSTM~\citep{Wang2015LearningNL} and achieves success in language inference tasks. 
% It adopts entity-by-entity attention in modeling soft alignment between entities and then pools a compact graph representation from entity representations.
% models with textual knowledge tbc
% \textbf{KV-Memory} (KVM)~\citep{Mihaylov2018KnowledgeableRE} encodes KG triples associated with $\mathbf{s}$ and uses a key-value memory module to enrich token representations from $\mathbf{s}$. This paradigm has proved to boost models' performance in reading comprehension tasks~\citep{hill2015goldilocks,zhang2018record}.

\section{Results and Discussions}
\label{sec:results}
% \vspace{-0.1cm}
In this section, we present the results of our models in comparison with baselines as well as methods on the leaderboards for both CommonsenseQA and OpenbookQA. We also provide analysis of models' components and characteristics.

% \begin{table}[h]
% \small
% \centering
% \begin{tabular}{lcc}
%   \hline
%   & {Dev} & {Test} \\ 
%   \hline
%   w/o KG$^{\dagger \ddagger}$ & $00.00 \pm 0.00$ & $00.00 \pm 0.00$   \\
%   RN & $00.00 \pm 0.00$ & $00.00 \pm 0.00$ \\
%   RGCN & $00.00 \pm 0.00$ & $00.00 \pm 0.00$ \\
%   KagNet & $00.00 \pm 0.00$ & $00.00 \pm 0.00$ \\
%   GconAttn & $00.00 \pm 0.00$ & $00.00 \pm 0.00$ \\
%   KV-Memory & $00.00 \pm 0.00$ & $00.00 \pm 0.00$ \\
%   \hline
%   MHGRN & $00.00 \pm 0.00$ & $00.00 \pm 0.00$  \\
%   \hline
% \end{tabular}
% \caption{Accuracy on OpenbookQA official split with \textsc{Roberta-Large} as text encoder.} \label{tab:results:obqa}
% \end{table}
% % \end{center}\subsection{Main Results}

\begin{table}
% \vspace{-0.1cm}
\centering
\scalebox{0.71}{
\begin{tabular}{@{}lcc@{}}
    \toprule
    \textbf{Methods} & Dev (\%) & Test (\%)\\
    \midrule
    {T5-3B\textsuperscript{\dag}~\cite{t5}}  & - & 83.20 \\
    {UnifiedQA\textsuperscript{\dag}~\cite{khashabi2020unifiedqa}}  & - & \textbf{87.20} \\
    % {BERT (w/o KG)\textsuperscript{\dag}}  & 60.4& 60.4 \\
    % {BERT Multi-Task (w/o KG)\textsuperscript{\dag}}  & - &  63.8 \\
    % GapQA\textsuperscript{\dag}~\citep{khot2019s} & - & 59.4~($\pm$1.30) \\
    \midrule
    {RoBERTa-Large (w/o KG)}  & 66.76~($\pm$1.14)& 64.80~($\pm$2.37) \\
    { + RGCN}  & 64.65~($\pm$1.96)& 62.45~($\pm$1.57) \\
    { + GconAttn}  & 64.30~($\pm$0.99)& 61.90~($\pm$2.44) \\
    % { + KVM}  & 69.40~($\pm$1.70)& 67.40~($\pm$0.75) \\
    { + RN (1-hop)} & 64.85~($\pm$1.11) & 63.65~($\pm$2.31) \\
    { + RN (2-hop)}  & 67.00~($\pm$0.71) & 65.20~($\pm$1.18)\\
    % { + MHGRN ($K=3$)}  & xx.xx~($\pm$x.xx) & xx.xx~($\pm$x.xx) \\
    { + MHGRN ($K=3$)}  & 68.10~($\pm$1.02) & \textbf{66.85}~($\pm$1.19) \\
    \midrule
    AristoRoBERTaV7\textsuperscript{\dag} & 79.2 & 77.8 \\
    {~+ MHGRN ($K=3$)} & 78.6 & \textbf{80.6} \\
    \bottomrule
\end{tabular}
}
% \vspace{-0.2cm}
\caption{\textbf{Performance comparison on OpenbookQA.} $\dagger$ indicates reported results on leaderboard. T5-3B is 8 times larger than our models. UnifiedQA is 30x larger.}
\label{tab:results:obqa}
\vspace{-0.2cm}
\end{table}

\vspace{-0.1cm}
\subsection{Main Results}
\label{subsec:main_results}
\vspace{-0.1cm}
% The results are presented in Table~\ref{tab:results:csqa} Table~\ref{tab:results:obqa}
% Table~\ref{tab:results:csqa_official}
% For CommonsenseQA~(Table~\ref{tab:results:csqa}), we first use data split of \citet{kagnet-emnlp19} to compare our models with our implemented baselines. Almost all KG-augmented models achieve performance gain over vanilla pre-trained LMs, demonstrating the value of external knowledge on this dataset. Additionally, we evaluate models on official Dev ~(Table~\ref{tab:results:csqa_official}) for fair comparison with other methods on leaderboard, where such patterns also apply. On both splits, MHGRN substantially surpasses all baselines regardless of the choice of text encoder, indicating our model's consistent superiority over existing KG-augmented methods. 
For CommonsenseQA~(Table~\ref{tab:results:csqa}), we first use the in-house data split~(IH) of \citet{kagnet-emnlp19}~(cf. Appendix~\ref{app:ds}) to compare our models with implemented baselines. This is \textbf{different from the official split} used in the leaderboard methods. Almost all KG-augmented models achieve performance gain over vanilla pre-trained LMs, demonstrating the value of external knowledge on this dataset. 
Additionally, we evaluate our MHGRN (with text encoder being \textsc{RoBERTa-Large}) on \textbf{official split}, OF ~(Table~\ref{tab:results:csqa_official}) for fair comparison with other methods on leaderboard, in both single-model setting and ensemble-model setting. 
With backbone being T5~\cite{t5}, UnifiedQA~\cite{khashabi2020unifiedqa} tops the leaderboard. Considering its training cost, we do not intend to compare our \textsc{RoBERTa}-based model with it. We achieve the best performances among other models.

For OpenbookQA~(Table~\ref{tab:results:obqa}), we use official split and build models with \textsc{RoBERTa-Large} as text encoder. MHGRN surpasses all implemented baselines, with an absolute increase of $\sim$2\% on Test. Also, as our approach is naturally compatible with the methods that utilize textual knowledge or extra data, because in our paradigm the encoding of textual statement and graph are structurally-decoupled~(Fig.~\ref{fig:overview}). To empirically show MHGRN can bring gain over textual-knowledge empowered systems, we replace our text encoder with AristoRoBERTaV7\footnote{\scriptsize\url{https://leaderboard.allenai.org/open_book_qa/submission/blcp1tu91i4gm0vf484g}}, and fine-tune our MHGRN upon OpenbookQA. Empirically, MHGRN continues to bring benefit to strong-performing textual-knowledge empowered systems. This indicates textual knowledge and structured knowledge can potentially be complementary.
% As we seek to evaluate the performance of models in terms of encoding external knowledge graphs,
% we did not compare with other models based on document retrieval and reading comprehension.
% Note that other submissions on OBQA leaderboard are usually fine-tuned on external question answering datasets or using large corpus via information retrieval.
% As we seek to systematically examine knowledge-aware reasoning methods  by reasoning over interpretable structures,
% we do not include comparisons with  models that are beyond the scope.
% Other fine-tuned PTLMs can be simply incorporated as well whenever they are available.

\subsection{Performance Analysis}
\label{subsec:analysis}

\begin{table}[t]
% \vspace{-0.1cm}
\centering
\scalebox{0.8}{
\begin{tabular}{@{}lcc@{}}
    \toprule
    \textbf{Methods} & IHdev-Acc. (\%) \\
    \midrule
    MHGRN ($K=3$) & \textbf{74.45}~($\pm$0.10)  \\
    {~- Type-specific transformation (\sectionref{subsec:method:arch})} & 73.16~($\pm$0.28) \\
    {~- Structured relational attention (\sectionref{subsec:method:att})} & 73.26~($\pm$0.31) \\
    {~- Relation type attention (\sectionref{subsec:method:att})} & 73.55~($\pm$0.68) \\
    {~- Node type attention (\sectionref{subsec:method:att})} & 73.92~($\pm$0.65) \\
    \bottomrule
\end{tabular}
}
% \vspace{-0.2cm}
\caption{\textbf{Ablation study on model components} (removing one component each time) using \textsc{Roberta-Large} as the text encoder. We report the IHdev accuracy on CommonsenseQA.}
\label{tab:results:ablation}
% \vspace{-0.0cm}
\end{table}

\noindent
\textbf{Ablation Study on Model Components.}
As shown in Table~\ref{tab:results:ablation}, disabling type-specific transformation results in $\sim1.3\%$ drop in performance, demonstrating the need for distinguishing node types for QA tasks. Our structured relational attention mechanism is also critical, with its two sub-components contributing almost equally.

\begin{figure}[h]
% \vspace{-0.2cm}
      \centering
        \includegraphics[width=0.8\linewidth]{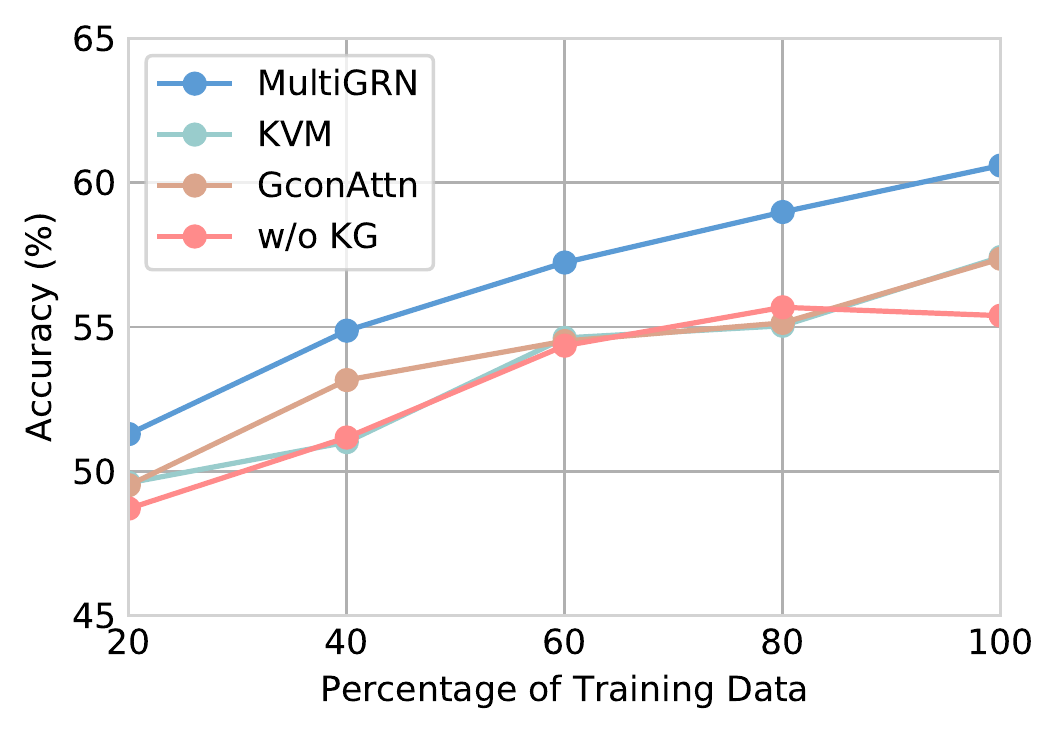}
        \vspace{-0.2cm}
      \caption{Performance change (accuracy in \%) w.r.t. the amounts of training data on CommonsenseQA IHTest set (same as \citet{kagnet-emnlp19}).}
      \label{fig:amount}
\vspace{-0.1cm}
\end{figure}

\smallskip
\noindent
\textbf{Impact of the Amount of Training Data.}
% \subsection{Impact of the Amount of Training Data}
We use different fractions of training data of CommonsenseQA and report results of fine-tuning text encoders alone and jointly training text encoder and graph encoder in Fig.~\ref{fig:amount}. Regardless of training data fraction, our model shows consistently more performance improvement over knowledge-agnostic fine-tuning compared with the other graph encoding methods, indicating MHGRN's complementary strengths to text encoders.
% and the gap grows as amount of training data grows, indicating our model's potential to model higher-order relations if given more data.

\begin{figure}[ht]
\vspace{-0.1cm}
      \centering
        \includegraphics[width=0.75\linewidth]{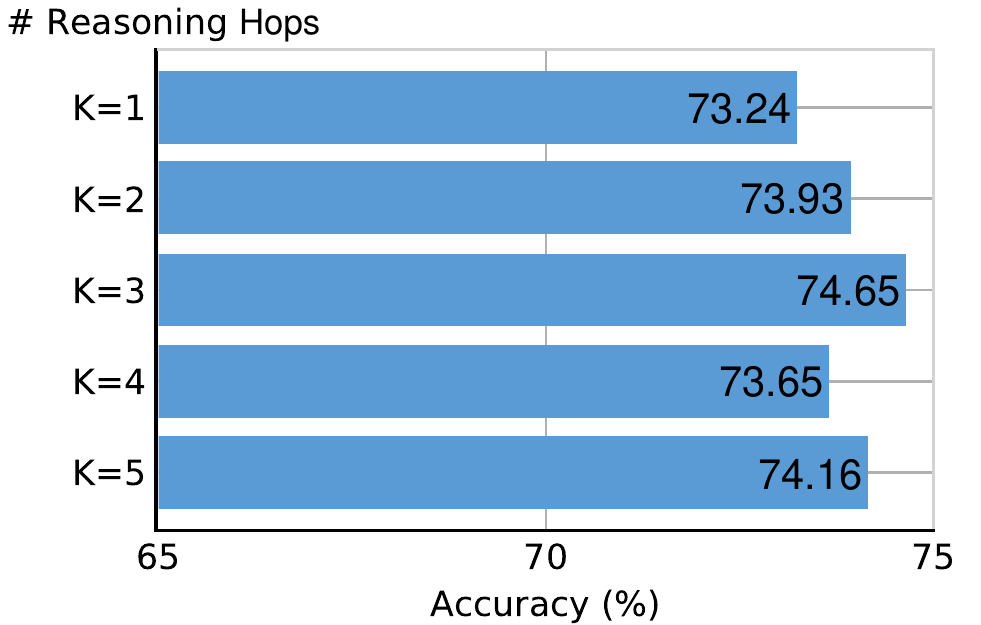}
        \vspace{-0.1cm}
      \caption{\textbf{Effect of $K$ in MHGRN}. We show IHDev accuracy of MHGRN on CommonsenseQA w.r.t. \# hops.}
      \label{fig:impact_k}
\vspace{-0.2cm}
\end{figure}

\smallskip
\noindent
\textbf{Impact of Number of Hops ($K$).}
We investigate the impact of hyperparameter $K$ for MHGRN by its performance on CommonsenseQA (Fig.~\ref{fig:impact_k}). The increase of $K$ continues to bring benefits until $K=4$. However, performance begins to drop when $K > 3$. This might be attributed to exponential noise in longer relational paths in knowledge graph.

% \smallskip
% \noindent
% \textbf{Study of Model Scalability.}
\subsection{Model Scalability}
\begin{figure}[t]
\vspace{-0.0cm}
      \centering
        \includegraphics[width=0.7\linewidth]{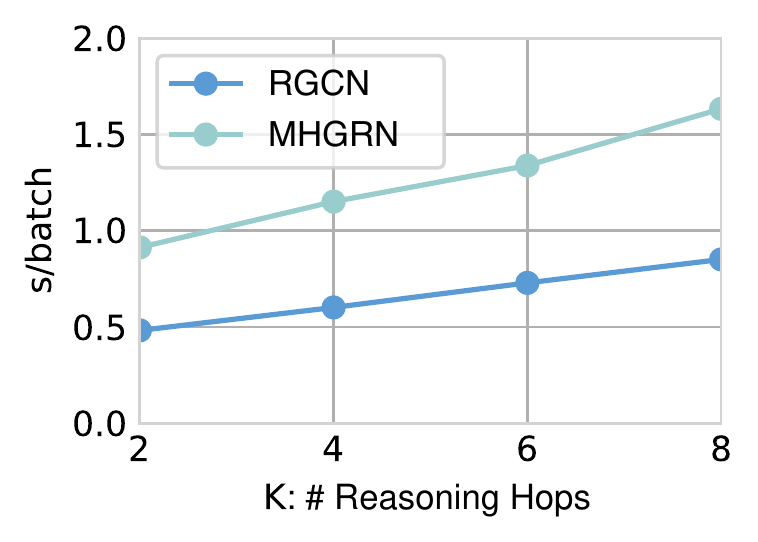}
    \vspace{-0.2cm}
      \caption{\textbf{Analysis of model scalability}. Comparison of per-batch training efficiency  w.r.t. \# hops $K$.}
      \label{fig:scale}
\vspace{-0.2cm}
\end{figure}

Fig.~\ref{fig:scale} presents the computation cost of MHGRN and RGCN (measured by training time). Both grow linearly w.r.t. $K$. Although the theoretical complexity of MultiRGN is $m$ times that of RGCN, the ratio of their empirical cost only approaches $2$, demonstrating that our model can be better parallelized.

% \smallskip
% \noindent
% \textbf{Case Study.}
\subsection{Model Interpretability}
\begin{figure}[h]
\vspace{-0.1cm}
      \centering
        \includegraphics[width=0.97\linewidth]{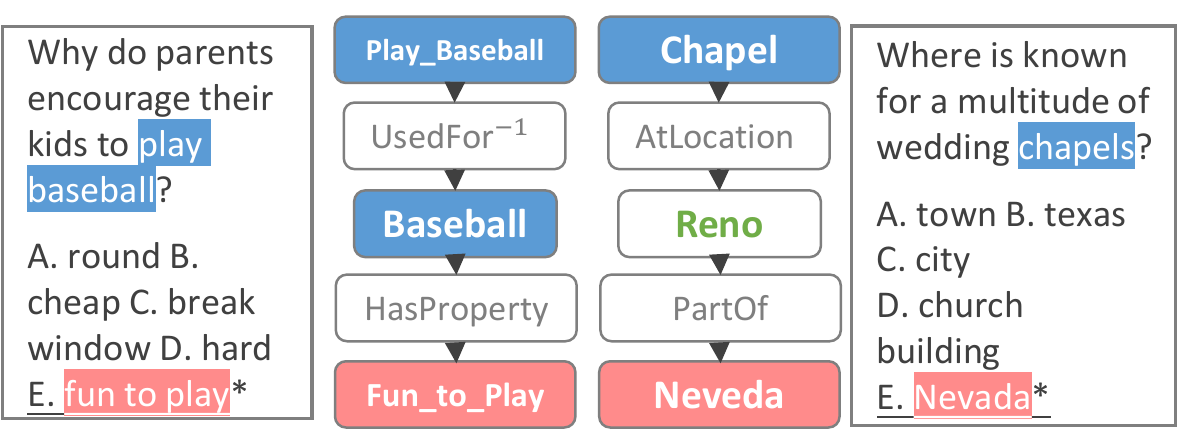}
        \vspace{-0.0cm}
      \caption{\textbf{Case study on model interpretability.} We present two sample questions from CommonsenseQA with the reasoning paths output by MHGRN.}
      \label{fig:case}
\vspace{-0.2cm}
\end{figure}

% \xiang{Here we make a figure to show 2-3 concrete examples of how one could understand the model behaviors via the model's output.
% \\
% (from Peifeng) Probably need to cover the cases where RGCN or LM fail. And also cases where different hops are leveraged.}
%\input{figures/case.tex}
We can analyze our model's reasoning process by decoding the reasoning path using the method described in \sectionref{subsec:method:learning}. Fig.~\ref{fig:case} shows two examples from CommonsenseQA, where our model correctly answers the questions and provides reasonable path evidences. In the example on the left, the model links question entities and answer entity in a chain to support reasoning, while the example on the right provides a case where our model leverage unmentioned entities to bridge the reasoning gap between question entity and answer entities, in a way that is coherent with the latent relation between \textsc{Chapel} and the desired answer in the question.
% Besides giving the correct answer, our model provides the reasoning path as \begin{small}$(\textsc{Chapel}\rightarrow \texttt{AtLocation}\rightarrow\texttt{PartOf}\rightarrow\textsc{Nevada})$\end{small}, which is coherent with the implied relation between \textsc{Chapel} and the desired answer in the question.

% \subsection{Compatibility with Other Methods}
% \input{tables/compatibility}
% Our approach is naturally compatible with the methods that utilize textual knowledge or extra data, because in our paradigm the encoding of textual statement and graph are structurally-decoupled~(Fig.~\ref{fig:overview}). We can directly take these QA system that deals with textual knowledge as our text encoder and leave the rest of our model architecture unchanged. To empirically show that our method can bring gain over textual-knowledge empowered systems, we replace our text encoder with AristoRoBERTaV7\footnote{\url{https://leaderboard.allenai.org/open_book_qa/submission/blcp1tu91i4gm0vf484g}}, and fine-tune our MHGRN upon OpenbookQA. Empirically, MHGRN continues to bring benefit to textual-knowledge empowered systems. This indicates textual knowledge and structured knowledge can potentially be complementary.

% \subsection{Sensitivity Analysis}
% \xiang{Present curves or tables for some major hyper-params. Can move to appendix.}

\section{Related Work}
% \xiang{1. Pls double check to include the list of papers I mentioned on Slack, and stress on discussing the differences to these work (or distinguishing our focuses from theirs). 2. Compress to 1.2 column at most.}
% \xiang{compress more, to ~1 column}
\vspace{-0.1cm}
\noindent
\textbf{Knowledge-Aware Methods for NLP~}
% Yang2017LeveragingKB, weissenborn2017dynamic, Mihaylov2018KnowledgeableRE, Annervaz2018LearningBD, bauer-etal-2018-commonsense, kagnet-emnlp19, li-clark-2015-answering, kundu-etal-2019-exploiting, Wang2018ImprovingNL
% Zhang2019ERNIEEL
% Pan2019ImprovingQA, ye2019align, zhang2018kg, li2019teaching, Banerjee_2019
% strategies Sun_2019
Various work have investigated the potential to empower NLP models with external knowledge. 
Many attempt to extract \textit{structured} knowledge, either in the form of nodes~\citep{Yang2017LeveragingKB, Wang2018ImprovingNL}, triples~\citep{weissenborn2017dynamic,Mihaylov2018KnowledgeableRE}, paths~\citep{bauer-etal-2018-commonsense, kundu-etal-2019-exploiting, kagnet-emnlp19}, or subgraphs~\citep{li-clark-2015-answering}, and encode them to augment textual understanding. 

Recent success of pre-trained LMs motivates many~\citep{Pan2019ImprovingQA, ye2019align, zhang2018kg, li2019teaching, Banerjee_2019} to probe LMs' potential as latent knowledge bases. This line of work turn to \textit{textual} knowledge (\eg~Wikipedia) to directly impart knowledge to pre-trained LMs. They generally fall into two paradigms: 1) Fine-tuning LMs on large-scale general-domain datasets (\eg~RACE~\citep{lai-etal-2017-race}) or on knowledge-rich text. 2) Providing LMs with evidence via information retrieval techniques. However, these models cannot provide explicit reasoning and evidence, thus hardly trustworthy. 
They are also subject to the availability of in-domain datasets and maximum input token of pre-trained LMs. 

\smallskip
\noindent
\textbf{Neural Graph Encoding~}
% \xiang{1. This part should be a brief summary of GNNs; including RNs and not just GCNs. 2. In particular, should discuss some recent efforts on extending GNNs to model higher-order structure in graphs, such as \url{https://arxiv.org/abs/1810.02244}, and stress on the differeces.}
Graph Attention Networks (GAT)~\citep{Velickovic2017GraphAN} incorporates attention mechanism in feature aggregation, RGCN~\citep{Schlichtkrull2017ModelingRD} proposes relational message passing which makes it applicable to multi-relational graphs. However they only perform single-hop message passing and cannot be interpreted at path level. Other work~\citep{AbuElHaija2019MixHopHG, Nikolentzos2019khopGN} aggregate for a node its $K$-hop neighbors based on node-wise distances, but they are designed for non-relational graphs. 
%Our MHGRN overcomes these issues. It works on relational graphs and is interpretable at path level.
MHGRN addresses these issues by reasoning on multi-relational graphs and being interpretable via maintaining paths as reasoning chains.

%\citet{2019arXiv191108936S} adopts multi-hop message passing in KG alignment task but relation types only act as regularization.% and are not considered. 

%Their proposed GATs are widely applied to \textit{non-relational} graphs. The aforementioned work all adopt single-hop message passing.

% \xiang{below introduction are too detailed. Just use 1-2 sentences to summarize the shared limitations of these methods or whether their focus is different.}
%There are also attempts to enhance the representation capacity of GNNs via higher-order message passing within a single layer.
%\citet{AbuElHaija2019MixHopHG} and \citet{Nikolentzos2019khopGN} aggregate for a node its $K$-hop neighbors based on node-wise distances. These methods are however not applicable to relational graphs as their message passing does not take into account edge types. \citet{2019arXiv191108936S} adopt multi-hop message passing in KG alignment task by designating hop-specific transformation. However, relation types are not included in message passing and are only used as regularization. 
% \citet{AbuElHaija2019MixHopHG} separately aggregate a node's $K$-hop neighbors based on distances (rings) and concatenate them to form node representation. \citet{Nikolentzos2019khopGN} perform within-ring and across-ring information propagation alternately with proper integration. 

\section{Conclusion}
We present a principled, scalable method, MHGRN, that can leverage general knowledge via multi-hop reasoning over interpretable structures (e.g. ConceptNet). 
The proposed MHGRN generalizes and combines the advantages of GNNs and path-based reasoning models. 
It explicitly performs multi-hop relational reasoning and is empirically shown to outperform existing methods with superior scalablility and interpretability.
% Our extensive experiments systematically compare MHGRN and other existing methods on knowledge-aware methods.
% Particularly, we achieve the state-of-the-art performance on the CommonsenseQA dataset.

%and performs higher-order relational reasoning, therefore bringing about efficiency, transparency and outstanding performances on two datasets.
% \appendix
% \input{8-appendix.tex}

\bibliography{acl20-MultiGRN}
\bibliographystyle{acl_natbib}

\clearpage
\appendix
\section{Merging Types of Relations in ConceptNet}
\label{app:merge}
\begin{table}[h]
\centering

\begin{tabular}{cc}
    \toprule  
    \textbf{Relation} & \textbf{Merged Relation}\\
    \midrule
    \text{AtLocation} & \multirow{2}{*}{\text{AtLocation}} \\
    \text{LocatedNear} & \\
    \midrule
    \text{Causes} & \multirow{3}{*}{\text{Causes}} \\
    \text{CausesDesire} & \\
    \text{*MotivatedByGoal} & \\
    \midrule
    \text{Antonym} & \multirow{2}{*}{\text{Antonym}} \\
    \text{DistinctFrom} & \\
    \midrule
    \text{HasSubevent} & \multirow{6}{*}{\text{HasSubevent}} \\
    \text{HasFirstSubevent} & \\    \text{HasLastSubevent} & \\
    \text{HasPrerequisite} & \\
    \text{Entails} & \\
    \text{MannerOf} & \\
    \midrule
    \text{IsA} & \multirow{3}{*}{\text{IsA}} \\
    \text{InstanceOf} & \\    
    \text{DefinedAs} & \\
    \midrule
    \text{PartOf} & \multirow{2}{*}{\text{PartOf}} \\
    \text{*HasA} & \\    
    \midrule
    \text{RelatedTo} & \multirow{3}{*}{\text{RelatedTo}} \\
    \text{SimilarTo} & \\    
    \text{Synonym} & \\ 
    \bottomrule 
\end{tabular}
\caption{Relations in ConceptNet that are being merged in pre-processing. *RelationX indicates the reverse relation of RelationX.}
\label{tab:relations}
\end{table}
We merge relations that are close in semantics as well as in the general usage of triple instances in \textit{ConceptNet}.

\section{Dataset Split Specifications}
\label{app:ds}

\begin{table}[h]
\centering
\scalebox{0.85}{
\begin{tabular}{lccc}
  \toprule
   & {Train} & {Dev} & {Test} \\ 
  \midrule
  CommonsenseQA (OF) & $9,741$ & $1,221$ & $1,140$\\
  CommonsenseQA (IH) & $8,500$ & $1,221$ & $1,241$\\
  OpenbookQA & $4,957$ & $500$ & $500$\\
  \bottomrule
\end{tabular}
}
\caption{Numbers of instances in different dataset splits.} 
\label{tab:ds_split}
\end{table}
% \end{center}\subsection{Main Results}
CommonsenseQA~\footnote{\url{https://www.tau-nlp.org/commonsenseqa}} and OpenbookQA~\footnote{\url{https://leaderboard.allenai.org/open_book_qa/submissions/public}} all have their leaderboards, with training and development set publicly available.
As the ground truth labels for CommonsenseQA are not readily available, for model analysis, we take 1,241 examples from official training examples as our in-house test examples and regard the remaining 8,500 ones as our in-house training examples (CommonsenseQA (IH)).

\section{Implementation Details} \label{app:details}

\begin{table}[h]
\centering
\scalebox{0.80}{
\begin{tabular}{lcc}
  \toprule
   & {CommonsenseQA} & {OpenbookQA} \\ 
  \midrule
  \textsc{Bert-Base} & $3\times10^{-5}$ & - \\
  \textsc{Bert-Large} & $2\times10^{-5}$ & - \\
  \textsc{RoBERTa-Large} & $1\times10^{-5}$ & $1\times10^{-5}$ \\
  \bottomrule
\end{tabular}
}
\caption{Learning rate for text encoders on different datasets.}
\label{tab:elr}
\end{table}
% \end{center}\subsection{Main Results}

\begin{table}[h]
\centering
\scalebox{0.80}{
\begin{tabular}{lcc}
  \toprule
   & {CommonsenseQA} & {OpenbookQA} \\ 
  \midrule
  RN & $3\times10^{-4}$ & $3\times10^{-4}$ \\
  RGCN & $1\times10^{-3}$ & $1\times10^{-3}$ \\
  GconAttn & $3\times10^{-4}$ & $1\times10^{-4}$ \\
  % KV-Memory & $1\times10^{-3}$ & $1\times10^{-3}$ \\
  MHGRN & $1\times10^{-3}$ & $1\times10^{-3}$ \\
  \bottomrule
\end{tabular}
}
\caption{Learning rate for graph encoders on different datasets.} 
\label{tab:dlr}
\end{table}
% \end{center}\subsection{Main Results}

\begin{table}[ht]
\centering
\scalebox{0.80}{
\begin{tabular}{p{40mm} p{10mm}}
  \toprule
  {} &  {\#Param}  \\ 
  \midrule
  RN & $399K$  \\
  RGCN & $365K$ \\
  {GconAttn} & $453K$ \\

  MHGRN & $544K$  \\
  \bottomrule
\end{tabular}
}
\caption{Numbers of parameters of different graph encoders.} 
\label{tab:num_param}
\end{table}

Our models are implemented in \textit{PyTorch}. We use \textit{cross-entropy} loss and \textit{RAdam}~\citep{liu2019variance} optimizer. We find it beneficial to use separate learning rates for the text encoder and the graph encoder. We tune learning rates for text encoders and graph encoders on two datasets.
We first fine-tune \textsc{RoBERTa-Large}, \textsc{Bert-Large}, \textsc{Bert-Base} on \textit{CommonsenseQA} and \textsc{RoBERTa-Large} on \textit{OpenbookQA} respectively, and choose a dataset-specific learning rate from $\{1\times10^{-5}, 2\times10^{-5}, 3\times10^{-5}, 6\times10^{-5}, 1\times10^{-4}\}$ for each text encoder, based on the best performance on development set, as listed in Table~\ref{tab:elr}. 
We report the performance of these fine-tuned text encoders and also adopt their dataset-specific optimal learning rates in joint training with graph encoders. 
For models that involve KG, the learning rate of their graph encoders
% ~\footnote{KV-memory actually encode triples to enhance token representations of $\bm{s}$ instead of encoding graph alone. But for simplicity, we also refer to it as a graph encoder.} 
are chosen from $\{1\times10^{-4}, 3\times10^{-4}, 1\times10^{-3}, 3\times10^{-3}\}$, based on their best development set performance with \textsc{RoBERTa-Large} as the text encoder. We report the optimal learning rates for graph encoders in Table~\ref{tab:dlr}.
In training, we set the maximum input sequence length to text encoders to $64$, batch size to $32$, and perform early stopping. AristoRoBERTaV7+MHGRN is the only exception. In order to host fair comparison, we follow AristoRoBERTaV7 and set the batch size to $16$, max input sequence length to $256$, and choose a decoder learning rate from $\{1\times10^{-3}, 2\times10^{-5}\}$.

For the input node features, we first use templates to turn knowledge triples in \textit{ConceptNet} into sentences and feed them into pre-trained \textsc{Bert-Large}, obtaining a sequence of token embeddings from the last layer of \textsc{Bert-Large} for each triple. For each entity in \textit{ConceptNet}, we perform mean pooling over the tokens of the entity's occurrences across all the sentences to form a $1024d$ vector as its corresponding node feature. We use this set of features for all our implemented models.

We use 2-layer RGCN and single-layer MHGRN across our experiments.

The numbers of parameter for each graph encoder are listed in Table~\ref{tab:num_param}.
% \section{Dynamic Programming Algorithm for Eq.~\ref{eq:model_message_pass}}
\section{Dynamic Programming Algorithm for Eq. 7}
\label{app:dp}
                                                                                    
To show that multi-hop message passing can be computed in linear time, we observe that Eq. 7 can be re-written in matrix form:
\begin{multline} \label{eq:model_matrix}
    \bm{Z}^k = (\bm{D}^{k})^{-1}\sum_{(r_1, \dots, r_k)\in \mathcal{R}^k} \beta(r_1, \dots, r_k, \bm{s}) \\ \cdot \bm{G}\bm{A}_{r_k} \cdots\bm{A}_{r_1} \bm{F} \bm{X} {\bm{W}_{r_1}^1}^\top\cdots{\bm{W}_{r_k}^k}^\top \\ \cdot {\bm{W}_{0}^{k+1}}^\top \cdots {\bm{W}_{0}^K}^\top \quad (1\le k \le K),
\end{multline}
where $\bm{G}=\text{diag}(\exp([g(\phi(v_1), \bm{s}), \dots, g(\phi(v_n),\\ \bm{s})])$ ($\bm{F}$ is similarly defined), $\bm{A}_r$ is the adjacency matrix for relation $r$ and $\bm{D}^{k}$ is defined as follows:
\begin{multline}
    \bm{D}^k = \text{diag}\Bigg( \sum_{(r_1, \dots, r_k)\in \mathcal{R}^k} \beta(r_1, \dots, r_k, \bm{s}) \\ \cdot \bm{G}\bm{A}_{r_k}\cdots\bm{A}_{r_1} \bm{F} \bm{X} \bm{1} \Bigg) \quad (1\le k \le K)
\end{multline}
Using this matrix formulation, we can compute Eq. 7 using dynamic programming:

% $\bm{D}= \text{diag}\left( \sum_{r_1=0}^m\cdots\sum_{r_k=0}^m \beta(r_1, \dots, r_k, \bm{s}) \cdot \text{diag}(\bm{g})\bm{A}_{r_k}\cdots\bm{A}_{r_1} \text{diag}(\bm{g}^\prime) \bm{1} \right)$

% \footnote{$\bm{D}$ is defined by the un-normalized term in Eq. \ref{eq:model_matrix} after replacing $\bm{X}{\bm{W}_{r_1}^1}^\top\cdots {\bm{W}_{r_k}^k}^\top$ with $\bm{1}$}

\begin{algorithm}
\begin{small}
    \caption{Dynamic programming algorithm for multi-hop message passing.} \label{alg:message_pass}
    \textbf{Input:} $\mathbf{s}, \bm{X}, \bm{A}_r (1 \le r \le m), \bm{W}_r^t(r\in \mathcal{R}, 1\le t \le k),\bm{F}, \bm{G}, \delta, \tau$ \\
    \textbf{Output:} $\bm{Z}$ \\
    \begin{algorithmic}[1]
        \STATE $\hat{\bm{W}}^{K} \gets \bm{I}$
        \FOR{$k \gets K-1\text{ to }1$}
            \STATE $\hat{\bm{W}}^{k} \gets \bm{W}_0^{k+1}\hat{\bm{W}}^{k+1}$
        \ENDFOR
        \FOR {$r\in \mathcal{R}$}
            \STATE $\bm{M}_r \gets \bm{F} \bm{X}$
        \ENDFOR
        \FOR{$k \gets 1\text{ to }K$}
            \IF {$k > 1$}
                \FOR {$r\in \mathcal{R}$}
                    \STATE $\bm{M}_r^\prime \gets e^{
                    \delta(r, \mathbf{s})} \bm{A}_r \sum_{r^\prime\in \mathcal{R}} e^{\tau (r^\prime, r, \mathbf{s})}  \bm{M}_{r^\prime}{\bm{W}^k_{r}}^\top$
                \ENDFOR
                \FOR {$r\in \mathcal{R}$}
                    \STATE $\bm{M}_r \gets \bm{M}_r^\prime$
                \ENDFOR
            \ELSE
                \FOR {$r\in \mathcal{R}$}
                    \STATE $\bm{M}_r\gets e^{ \delta(r, \mathbf{s})} \cdot \bm{A}_r \bm{M}_{r}{\bm{W}^k_{r}}^\top$
                \ENDFOR
            \ENDIF
            \STATE $\bm{Z}^k \gets \bm{G} \sum_{r\in \mathcal{R}} \bm{M}_r \hat{\bm{W}}^k$
        \ENDFOR
        
        \STATE Replace $\bm{W}_{r}^t (0 \le r \le m, 1\le t \le k)$ with identity matrices and $\bm{X}$ with $\bm{1}$ and re-run line 1 - line 19 to compute $\bm{d}^1,\dots, \bm{d}^K$
        \FOR{$k \gets 1\text{ to }K$}
            \STATE $\bm{Z}^k \gets (\text{diag}(\bm{d}^k))^{-1} \bm{Z}^k$
        \ENDFOR
        \RETURN $\bm{Z}^1, \bm{Z}^2, \dots, \bm{Z}^k$
    \end{algorithmic}
\end{small}
\end{algorithm}

\section{Formal Definition of $K$-hop RN} \label{app:rn}
\begin{definition}
     ($K$-hop Relation Network) A multi-hop relation network is a function that maps a multi-relational graph to a fixed size vector:
    \begin{multline}
    \begin{small}\text{KHopRN}(\mathcal{G}; \tilde{\bm{W}}, \tilde{\bm{E}}, \tilde{\bm{H}}) = \sum_{k=1}^K \sum_{\substack{(j, r_1, \dots, r_k, i)\in \Phi_k \\ j\in \mathcal{Q} ~~i\in \mathcal{A}}}\end{small} \\ 
    \begin{small}\tilde{\beta}(j, r_1,\dots, r_k, i) \cdot \tilde{\bm{W}} ( \tilde{\bm{h}}_j \oplus (\tilde{\bm{e}}_{r_1} \circ \cdots \circ \tilde{\bm{e}}_{r_k}) \oplus \tilde{\bm{h}}_i ),\end{small}
\end{multline}
where $\circ$ denotes element-wise product and \begin{small}$\tilde{\beta}(\cdots)= 1 / (K|\mathcal{A}|\cdot |\{(j^\prime, \dots, i)\in \mathcal{G} \mid j^\prime \in \mathcal{Q} \}|)$\end{small} defines the pooling weights.
\end{definition}

\section{Expressing K-hop RN with MultiGRN} \label{app:proof}

\begin{theorem}
    Given any \begin{small}$\tilde{\bm{W}}, \tilde{\bm{E}}, \tilde{\bm{H}}$\end{small}, there exists a parameter setting such that the output of the model becomes \begin{small}$\text{KHopRN}(\mathcal{G}; \tilde{\bm{W}}, \tilde{\bm{E}}, \tilde{\bm{H}})$\end{small} for arbitrary $\mathcal{G}$.
\end{theorem} 

\textit{Proof.} Suppose $\tilde{\bm{W}}=[\tilde{\bm{W}}_1, \tilde{\bm{W}}_2, \tilde{\bm{W}}_3]$, where $\tilde{\bm{W}}_1, \tilde{\bm{W}}_3\in \mathbb{R}^{d_3\times d_1}, \tilde{\bm{W}}_2\in \mathbb{R}^{d_3\times d_2}$. For \textit{MHRGN}, we set the parameters as follows: $\bm{H} = \tilde{\bm{H}}, \bm{U}_{*} = [\bm{I}; \bm{0}]\in \mathbb{R}^{(d_1+d_2)\times d_1}, \bm{b}_{*}=[\bm{0}, \bm{1}]^\top\in \mathbb{R}^{d_1+d_2}, \bm{W}_r^t=\text{diag}(\bm{1} \oplus \tilde{\bm{e}}_r)\in \mathbb{R}^{(d_1+d_2)\times (d_1+d_2)} (r\in \mathcal{R}, 1\le t \le K), \bm{V}=\tilde{\bm{W}}_3\in \mathbb{R}^{d_3\times d_1}, \bm{V}^\prime=[\tilde{\bm{W}}_1, \tilde{\bm{W}}_2]\in \mathbb{R}^{d_3\times (d_1+d_2)}$. We disable the relation type attention module and enable message passing only from $\mathcal{Q}$ to $\mathcal{A}$. By further choosing $\sigma$ as the identity function and performing pooling over $\mathcal{A}$, we observe that the output of \textit{MultiGRN} becomes:
\begin{equation}
\begin{split}
    &\quad \frac{1}{|\mathcal{A}|}\sum_{i\in \mathcal{A}} \bm{h}_i^\prime \\
    &= \frac{1}{|\mathcal{A}|}\left(\bm{V}\bm{h}_i + \bm{V}^\prime \bm{z}_i \right) \\
    &= \frac{1}{K|\mathcal{A}|}\sum_{k=1}^K\left(\bm{V}\bm{h}_i + \bm{V}^\prime \bm{z}_i^k \right) \\
    &= \sum_{k=1}^K \sum_{\substack{(j, r_1, \dots, r_k, i)\in \Phi_k \\ j\in \mathcal{Q},~i\in \mathcal{A}}} \tilde{\beta}(j, r_1, \dots, r_k, i) \Big(\bm{V}\bm{h}_i + \\ &\quad  \bm{V}^\prime \bm{W}_{r_k}^k \cdots \bm{W}_{r_1}^1 \bm{x}_j \Big) \\
    &= \sum_{k=1}^K \sum_{\substack{(j, r_1, \dots, r_k, i)\in \Phi_k \\ j\in \mathcal{Q}}} \tilde{\beta}(\cdots) \Big(\bm{V}\bm{h}_i + \bm{V}^\prime \bm{W}_{r_k}^k  \\ &\quad  \cdots \bm{W}_{r_1}^1 \bm{U}_{\phi(j)} \bm{h}_j + \bm{V}^\prime \bm{W}_{r_k}^k \cdots \bm{W}_{r_1}^1 \bm{b}_{\phi(j)} \Big) \\
    &=\sum_{k=1}^K \sum_{\substack{(j, r_1, \dots, r_k, i)\in \Phi_k \\ j\in \mathcal{Q},~i\in \mathcal{A}}} \tilde{\beta}(\cdots) \Big(\tilde{\bm{W}}_3\bm{h}_i + \tilde{\bm{W}}_1 \bm{h}_j  \\ &\quad + \tilde{\bm{W}}_2 (\tilde{\bm{e}_{r_1}}\circ \cdots \circ \tilde{\bm{e}_{r_k}}) \Big) \\
    &= \sum_{k=1}^K \sum_{\substack{(j, r_1, \dots, r_k, i)\in \Phi_k \\ j\in \mathcal{Q},~i\in \mathcal{A}}} \tilde{\beta}(\cdots) \tilde{\bm{W}}\Big(\tilde{\bm{h}}_j\oplus  \\ &\quad (\tilde{\bm{e}}_{r_1} \circ \cdots \circ \tilde{\bm{e}}_{r_k}) \oplus \tilde{\bm{h}}_i \Big) \\
    &= \textit{RN}(\mathcal{G}; \tilde{\bm{W}}, \tilde{\bm{E}}, \tilde{\bm{H}})
\end{split}
\end{equation}

\end{document}